\documentclass[11pt,a4paper]{article}
\usepackage{times,latexsym}
\usepackage{url}
\usepackage[T1]{fontenc}
\usepackage{tabu}
\usepackage{multirow}
\usepackage{caption}
\usepackage{booktabs}
\usepackage{amsmath}
\usepackage{geometry}  
\usepackage{lipsum}
\usepackage{xspace}
\usepackage{arydshln}
\usepackage{rotate}
\usepackage{makecell}
\usepackage{float}

\usepackage{graphicx}
\usepackage[acceptedWithA]{tacl2021v1}
\usepackage{subcaption}

\usepackage{xspace,mfirstuc,tabulary}
\usepackage{fontawesome}

\newcommand{\spot}{\mbox{\textsc{Spirit LM}}\xspace}
\newcommand{\llamatwo}{\mbox{\textsc{Llama 2}}\xspace}
\newcommand{\whisper}{\textsc{Whisper}\xspace}
\newcommand{\whispermedium}{\textsc{Whisper-Medium}\xspace}
\newcommand{\mmstts}{\textsc{MMS-TTS}\xspace}
\newcommand{\holisticbias}{\textsc{HolisticBias}\xspace}
\newcommand{\etox}{\textsc{ETOX}\xspace}
\newcommand{\asretox}{\textsc{ASR-ETOX}\xspace}
\newcommand{\mutox}{\textsc{MuTox}\xspace}
\newcommand{\emov}{\textsc{EmoV}\xspace}
\newcommand{\expresso}{\textsc{Expresso}\xspace}
\newcommand{\expressoasr}{\textsc{Expresso-ASR}\xspace}
\newcommand{\expressoread}{\textsc{Expresso-read}\xspace}
\newcommand{\base}{\textsc{Base}\xspace}
\newcommand{\expressive}{\textsc{Expressive}\xspace}

\newcommand{\expressiveVone}{\spot{} \textsc{Expressive}\xspace}
\newcommand{\spotexpressive}{\spot{} \expressive}
\newcommand{\spotbase}{\spot{} \base}

\newcommand{\sentimentbenchmark}{ \textsc{Speech-Text Sentiment Preservation} benchmark\xspace}

\newcommand{\sentimentbenchmarkSHORT}{\textsc{STSP}\xspace}

\newcommand{\ben}[1]{{\color{blue} [Ben: #1]}}

\newcommand{\dpx}[1]{}
\newcommand{\tuanh}[1]{}

\newif\iftaclinstructions
\taclinstructionsfalse 
\iftaclinstructions
\renewcommand{\confidential}{}
\renewcommand{\anonsubtext}{(No author info supplied here, for consistency with
TACL-submission anonymization requirements)}
\newcommand{\instr}
\fi

\iftaclpubformat  

\else

\fi

\title{\spot{}: Interleaved Spoken and Written Language Model}

\makeatletter
\def\thanks#1{\protected@xdef\@thanks{\@thanks
        \protect\footnotetext{#1}}}
\makeatother

\author{
    Tu Anh Nguyen\thanks{$^{a,b,c}$ Equally contributed as co-first, co-second and co-last authors, resp.}$^{a,+,\dagger}$, Benjamin Muller$^{a,+}$, Bokai Yu$^{a,+}$, Marta R. Costa-jussa$^{b,+}$, \\
    \textbf{
    Maha Elbayad$^{b,+}$,
    Sravya Popuri$^{b,+}$,
    Christophe Ropers$^{b,+}$,
    Paul-Ambroise Duquenne$^{b,+,\dagger}$,
    }\\
    \textbf{
    Robin Algayres$^{b,\ddagger}$,
    Ruslan Mavlyutov$^{b,+}$,
    Itai Gat$^{b,+}$,
    Mary Williamson$^{b,+}$,
    }\\
    \textbf{
    Gabriel Synnaeve$^{c,+}$, Juan Pino$^{c,+}$, Benoît Sagot$^{c,\dagger}$, Emmanuel Dupoux$^{c,+,\ddagger}$
    }\\
    \\
    $^+$ Meta AI,
    $^\dagger$ Inria, Paris,
    $^\ddagger$ EHESS, ENS-PSL, CNRS, Paris
  \\
  \texttt{\{ntuanh, benjaminmuller, bokai, dpx\}@meta.com}
}

\date{}

\begin{document}
\maketitle
\begin{abstract}

\dpx{alternate}
    
     We introduce \spot{}, a  foundation multimodal language model that freely mixes text and speech. 
     Our model is based on a 7B pretrained text language model that we extend to the speech modality by continuously training it on text and speech units. Speech and text sequences are concatenated as a single stream of tokens, and trained with a word-level \textit{interleaving} method using a small automatically-curated speech-text parallel corpus. 
     \spot comes in two versions: a \textsc{Base} version that uses speech phonetic units (HuBERT) and an \textsc{Expressive} version that models expressivity using pitch and style units in addition to the phonetic units. For both versions, the text is encoded with subword BPE tokens.  The resulting model displays both the semantic abilities of text models and the expressive abilities of speech models. Additionally, we demonstrate that \spot{} can learn new tasks in a few-shot fashion across modalities (i.e. ASR, TTS, Speech Classification). 
     We make available model weights and inference code
     \footnote{\label{demopage}Generation samples can be found at: \url{https://speechbot.github.io/spiritlm} 
     }\footnote{\label{githubpage}Inference code and models are available at: \url{https://github.com/facebookresearch/spiritlm}}
     .


\end{abstract}
\vspace{-1.5em}
\section{Introduction}\label{sec:intro}
\vspace{-0.5em}
Prompting Large Language Models (LLMs) has become a standard in Natural Language Processing (NLP) since the release of GPT-3 \citep{brown2020gpt3}. 
Scaling language models to billions of parameters with massive datasets helps to achieve general-purpose language understanding and generation. Additionally, large-scale language models can solve new tasks by providing the model with a few examples through in-context few-shot learning.
Since then, a number of LLMs have been developed \cite{chowdhery2022palm, hoffmann2022training, zhang2022opt, touvron2023llama}.
Notably, LLaMA \citep{touvron2023llama} showed that smaller LLMs can achieve very good performance when training longer on more data using optimal-compute scaling laws \cite{kaplan2020scaling}, making LLMs more accessible for NLP research.

  
  

Speech Language Models (SpeechLMs), i.e. language models trained directly on speech, have been introduced \citep{gslm, algayres2023tgslm, borsos2023audiolm} and have recently become an active field of research 
\citep{Wang2023VallE, nguyen2023dgslm, hassid2023textually, rubenstein2023audiopalm}.
These models are either trained on speech-only datasets or datasets of specific tasks, e.g. Text-To-Speech (TTS), Automatic Speech Recognition (ASR) or Translation, making the LMs focus on certain modality or tasks and potentially loose their generalization capabilities.

Given the increasing quality of text-only LLMs \citep{brown2020gpt3,touvron2023llama2}, one successful approach to generate speech has been to build pipelines that first transcribe input speech with ASR,  then generate text using a text-only LLM and finally synthesize the generated text into speech with TTS. However, with such pipelines, modeling and generating expressive speech is constrained out of the language model, leading to poor generation from an expressive point of view.

\begin{figure*}[h]
\begingroup
\renewcommand{\arraystretch}{0.8} 
\begin{tabular}{lll}
a.&b.&c.\\
\raisebox{1.1em}{\includegraphics[width=0.16\linewidth]{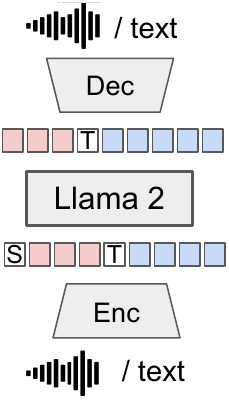}}&
\raisebox{1.8em}{\includegraphics[width=0.50\linewidth]{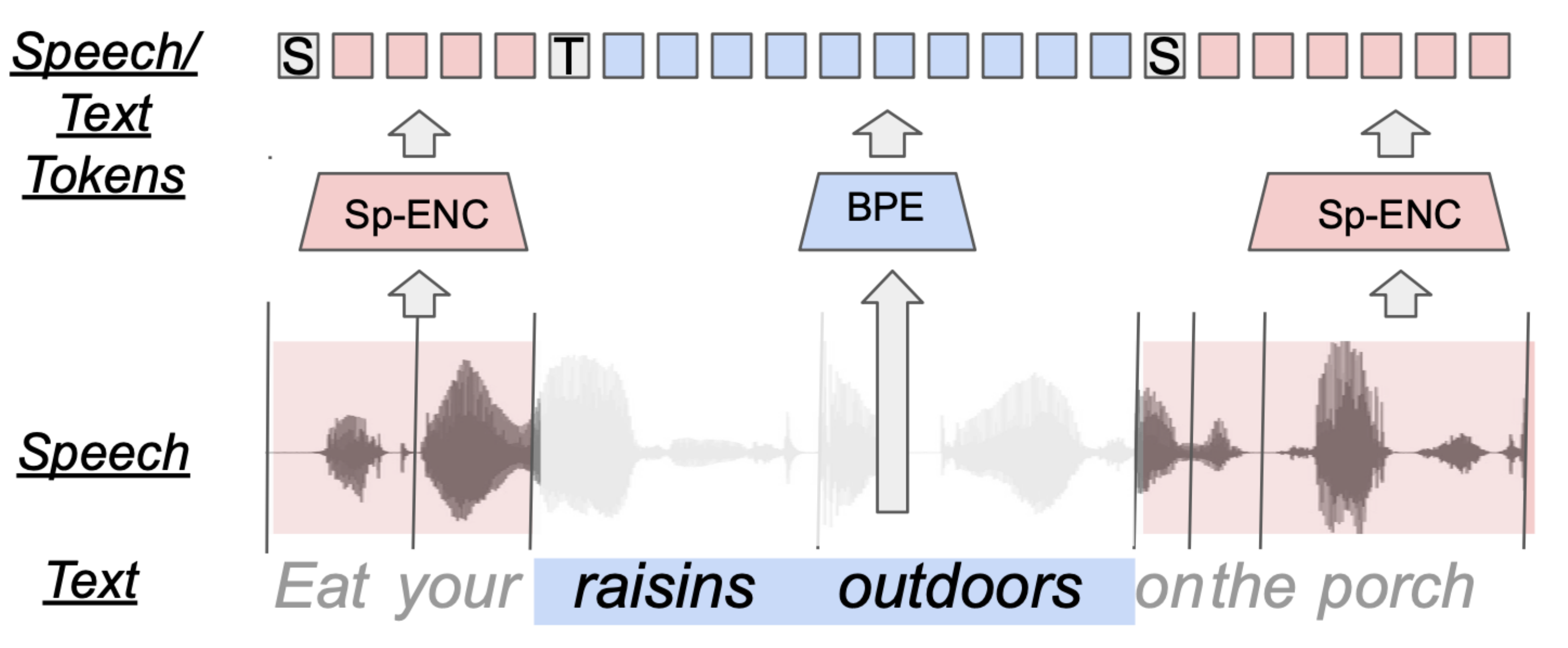}}&
\includegraphics[width=0.27\linewidth]{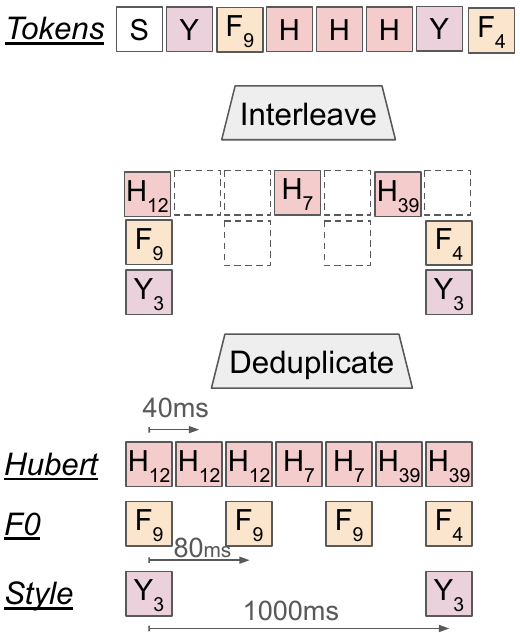}\\
\end{tabular}  
\endgroup
    \caption{\textbf{a.~The \spot{} architecture.} A 
    language model trained with next token prediction; tokens are derived from speech or text with an encoder, and rendered back in their original modality with a decoder. \spot{} models are trained on a mix of text-only sequences, speech-only sequences, and \textit{interleaved} speech-text sequences. 
    \textbf{b.~Speech-text interleaving scheme.} Speech is encoded into tokens (pink) using clusterized speech units (Hubert, Pitch, or Style tokens), and text (blue) using BPE. We use special tokens \textsc{[Text]} to prefix text and \textsc{[Speech]}  for speech tokens. During training, a change of modality is randomly triggered at word boundaries in aligned speech-text corpora. Speech tokens are deduplicated and interleaved with text tokens at the modality change boundary. \textbf{c.~Expressive Speech tokens.} For \spotexpressive, pitch tokens and style tokens are interleaved after deduplication.}
  \label{fig:boat1}
\end{figure*}

In this work, we aim to combine the generative abilities and pretrained knowledge of text LLMs with the expressive capacities of speech-language models. We show that LLMs trained on interleaved speech and text can learn speech and text cross-modally and are able to generate language content in either modality.
We evaluate the models with comprehension tasks in both speech and text, and extend few-shot prompting to speech-text tasks such as ASR, TTS or Speech Classification.
We further extend the phonetic speech tokens with expressive tokens that capture the pitch and style of the speech, and evaluate the models with newly introduced sentiment modeling tasks. Our contributions are the following:
(i) We introduce \spot{}, a single language model that can generate both speech and text. \spot is based on continuously pretraining \llamatwo with \textit{interleaving} speech and text data. (ii) Similarly to text LLMs, we find that \spot can learn new tasks in the few-shot setting in text, speech and in the cross-modal setting (i.e. speech to text and text to speech). (iii) To evaluate the expressive abilities of generative models, we introduce the \sentimentbenchmark (noted \sentimentbenchmarkSHORT) that measures how well generative models preserve the sentiment of the prompt within and across modalities for both spoken and written utterances\footnote{\label{stsppage}\sentimentbenchmarkSHORT evaluation code is available at: \url{https://github.com/facebookresearch/spiritlm/tree/main/spiritlm/eval}}. (iv) We propose an expressive version of \spot (\spotexpressive). Using \sentimentbenchmarkSHORT, we show that \spot is the first LM that can preserve the sentiment of text and speech prompts both within and across modalities. (v) Finally, we quantify the potential added toxic content in the generation of our model for both speech and text. As all pretrained base models \citep{stochastic_parrots,Solaiman}, \spot can generate harmful content. For these reasons, all user-facing applications using our work should integrate the necessary red-teaming work and implement safety instruction-tuning to meet safety standards \citep{touvron2023llama2}.\footnote{We point to the safety tuning done in \textsc{Llama 2-chat} for best practice references.} 

\begin{table*}[t]
\centering
\footnotesize
\begin{tabular}{p{1.5cm}p{5.9cm}p{7.2cm}}
     \toprule
     \multirow{2.5}{*}{
     \textbf{Inference}}  &  
     \multicolumn{2}{c}{\textbf{\spot generations}}\\
     \cmidrule{2-3}
    &  \textbf{Prompt} &  \textbf{Generation}\\
     \midrule
     \multicolumn{3}{c}{\textbf{\spotbase}}\\
    S$\rightarrow$S & \textsc{[Speech]}[Hu34][Hu301][Hu280]..[Hu34]& [Hu28][Hu41][Hu123]..[Hu254] \\
& \faVolumeUp \textit{~a b c d e} & \faVolumeUp \textit{~f g h i j k l m n o p q r c s t u v}\\
    T$\rightarrow$S 
   & \textsc{[Text]}The largest country in the world is & \textsc{[Speech]}[Hu34][Hu20][Hu340]..[Hu489] \\
   &  & \faVolumeUp  \textit{~Russia. It has about 150 millions inhabitants..}  \\
   T$\rightarrow$S & \textsc{[Text]}Here's a story about a llama that & \textsc{[Speech]}[Hu12][Hu41][Hu123]..[Hu254] \\
   & can speak:  &  \faVolumeUp  \textit{~This little llama had a friend named dobby...}\\
 S$\rightarrow$T
   & \textsc{[Speech]}[Hu34][Hu71][Hu405]..[Hu34] \faVolumeUp  \textit{~Yellowstone national park is an american national park located in}  & \textsc{[Text]} the northwest corner of Wyoming. It is located in the Greater Yellowstone area... 
   \\
    S$\rightarrow$T 
   & \textsc{[Speech]}[Hu34][Hu301][Hu280]..[Hu34]	
 & \textsc{[Text]} 6 7 8 9 10... \\
   &  \faVolumeUp~  \textit{one two three four five} &  \\
    \multicolumn{3}{c}{\textbf{\spotexpressive}}\\
S $\rightarrow$T
& \textsc{[Speech]}[St3][Pi0][Hu34][Hu103]..[Hu22] \faVolumeUp \textit{~Are you really going to do that <Angry>}  & \textsc{[Text]} he said in a voice that was almost a scream i'm afraid\\
S $\rightarrow$T
& \textsc{[Speech]}[St5][Pi5][Hu34][Hu409]..[Hu24] \faVolumeUp \textit{~Are you really going to do that <Disbelief>} & \textsc{[Text]} she said turning her head quickly and putting out her hand for the glasses\\
T$\rightarrow$S 
   & \textsc{[Text]}I am so deeply saddened & \textsc{[Speech]}[Hu34][St2][Pi9][Hu371][Hu20]..[Hu89] \\
   &  & \faVolumeUp  \textit{~...this moment is very very hard to me... <Sad>} \\
T$\rightarrow$S 
   & \textsc{[Text]}Your actions have made me incredibly angry & \makecell[tl]{\textsc{[Speech]}[Hu37][St1][Pi3][Hu38][Hu111]..[Hu98]\\ \faVolumeUp\textit{~So what you think you could talk about it to me <Angry>}} \\
   

\bottomrule
\end{tabular}
\caption{\spot generations with text (T) or speech (S) prompt and elicited to generate text (marked with special token \textsc{[Text]}) or speech (marked with special token \textsc{[Speech]}). We report the transcripted speech examples under the speech sequence indicated with \faVolumeUp \ and < > (e.g., <Angry>) is appended when the speech is presented with the associated emotion. \spot models are \mbox{Llama-2 7B} models \citep{touvron2023llama} fine-tuned with text (BPE) and 
speech tokens where Hubert token (cf.\S~\ref{sec:spotbase}) is denoted as [Hu], while [Pi] and [St], used exclusively in \spotexpressive (cf.\S~\ref{sec:spotexpressive}), represent the Pitch token and the Style token, respectively. \spot models enable semantically consistent multimodal generations, few-shot learning for text and speech tasks, cross-modal inference (text to speech and speech to text) and expressive generations. 
The samples can be found on the demo website\textsuperscript{\ref{demopage}}.
}
\label{tab:generations_tacl}
\end{table*}

\section{Related Work}\label{sec:related}
\paragraph{Textless NLP} Recent progress in Self-Supervised Speech Representation Learning (SSL) \citep{baevski2020w2v2, hsu2021hubert, chen2022wavlm, chung2021w2vbert}
has made it possible to learn from raw audio speech representations that are good for a variety of downstream tasks (\citealp{yang2021superb}). In addition, these methods can be used to derive discrete tokens that operate as a kind of pseudo-text and can be used to learn a language model from raw audio \citep{gslm} which is able to capture both the linguistic content and the prosody \cite{kharitonov2022pgslm}, giving rise to a host of applications: emotion conversion \citep{kreuk2022emotionconversion}, dialogue generation \citep{nguyen2023dgslm}, speech classification \citep{chang2023speechpromptv2}. Even though these models are good at capturing expressivity, they trail text models in capturing semantics when trained with comparable amounts of data (see \citealp{nguyen2020zero, nguyen2023dgslm}).
In this work, we use phonetic speech tokens extracted from HuBERT \citep{hsu2021hubert}, possibly combined with pitch and style tokens (as in \citealp{kharitonov2022pgslm}), and supplement the model with textual bpe-units. 

\paragraph{Speech and Speech+Text LMs} There has been an increasing number of SpeechLMs since GSLM \citep{gslm}.
AudioLM \citep{borsos2023audiolm} utilizes two types of discrete speech tokens with phonetic tokens\footnote{this is mentioned as \textit{semantic tokens} in their work, but we call \textit{phonetic tokens} as they capture phonetic rather than semantic information from the speech} \citep{chung2021w2vbert}, 
 and acoustic tokens 
\citep{zeghidour2021soundstream} to capture phonetic and acoustic information from speech respectively. 
Vall-E \citep{Wang2023VallE} models speech with acoustic tokens (Encodec, \citealp{defossez2022highfi}) and perform TTS task by translating phonemes to tokens using an autoregressive LM.
\citet{hassid2023textually} found that fine-tuning pre-trained TextLMs helps boost the performance of SpeechLMs.
SpeechGPT \citep{zhang2023speechgpt} further fine-tune speechLMs on cross-modal tasks (ASR, TTS) and chain-of-modality Question-Answering (QA) task. 
Similar to SpeechGPT, Spectron \citep{nachmani2023spoken} utilizes text as a proxy for spoken QA and speech continuation tasks. 
Unlike previous work, they represent speech using a spectrogram with pre-trained speech encoder from \citealp{zhang2023usm}. 
In the same spirit, \citet{fathullah2023generalpurpose} adapted \llamatwo  for speech generation tasks. 
AudioPALM \cite{rubenstein2023audiopalm} and VioLA \cite{wang2023viola} both train autoregressive language models on text and speech in a multi-task fashion. 
Most recently, VoxtLM \citep{maiti2023voxtlm} and SUTLM \citep{chou2023toward} jointly trained speech and text LMs on ASR, TTS, and speech/text continuation tasks.
Our work is similar to \citet{chou2023toward} in the training tasks but with the capacity of performing cross-modal generation and expressive speech and text generation.
We also study larger models and evaluate their zero-shot and in-context learning capabilities.



\begin{table}[t]
\setlength{\tabcolsep}{2pt}
\centering\small
\resizebox{0.5\textwidth}{!}{
\begin{tabu}{l l c c c c c}
    \toprule
    ~ & 
    & \multirow{2}{*}{\textbf{Hours}} & \multicolumn{2}{c}{\textbf{N Tokens}} & \multirow{2}{*}{\textbf{P Samp.}} & \multirow{2}{*}{\textbf{Epochs}}  \\
    \cline{4-5}
    ~ & ~ & ~ & Speech & Text & ~ & ~ \\
    \midrule
\multirow{1}{*}{Speech-only} &  & 458K & 28.2B & ~ & 33.3\% & 1.24  \\ 
    
 \multirow{1}{*}{Speech+Text} &  & 111K & 7.0B & 1.4B & 33.3\% & 3.81  \\
       

    \multirow{1}{*}{Text-only} &  & ~ & ~ & 307B & 33.3\% & 0.11  \\ 
    
    \bottomrule
\end{tabu}
}
\caption{\textbf{Statistics of training data.} P Samp. is the Sampling Proportion of each subset for a training batch. Epochs is the number of epochs seen for each subset after 100K training steps or equivalently 100B tokens. 
}
\label{tab:data}
\vspace{-1.2em}
\end{table}

\vspace{-0.5em}

\section{\spot Training Recipe}\label{sec:methods}
\vspace{-0.5em}

\spot models are based on continuously pretraining a text-pretrained language model on a combination of text and speech (Figure \ref{fig:boat1}.a).
Following \citealp{hassid2023textually}, we continuously pretrain \llamatwo \citep{touvron2023llama2} using a collection of text-only datasets, speech-only datasets and aligned speech+text datasets fed to the model with \textit{interleaving}. 

\spot comes in two versions: \base and \expressive. \spotbase models speech using HuBERT  tokens \citep{hsu2021hubert} while \spotexpressive uses the concatenation of HuBERT, pitch and style tokens. 

\subsection{\spotbase} \label{sec:spotbase}
The \spotbase model is based on the 7B version of \llamatwo trained on Text-only, Speech-only, and aligned Speech+Text datasets. 

\paragraph{Speech Encoder} We use the same HuBERT model as in TWIST \citep{hassid2023textually}, which is trained on a mixture of datasets: Multilingual LibriSpeech \citep{pratap20mls}, Vox Populi \citep{wang-etal-2021-voxpopuli}, Common Voice \citep{ardila-etal-2020-commonvoice}, Spotify \citep{clifton-etal-2020-spotify}, and Fisher \citep{Cieri2004TheFC}. 
and obtain a vocabulary of 501 phonetic speech tokens.

\begin{table}[t]
\setlength{\tabcolsep}{2pt}
\centering\footnotesize{
\resizebox{0.49\textwidth}{!}{
\begin{tabular}{lccccccc}
\toprule
\textbf{Model}& \bf{\#shots} & \multicolumn{4}{c}{\bf Accuracy $\uparrow$} 
\\
&&\textbf{T$\rightarrow$T} & \textbf{T$\rightarrow$S} & \textbf{S$\rightarrow$S} & \textbf{ S$\rightarrow$T} &\it Avg\\
\midrule



\spotbase & 0    & 0.65  & 0.33 & 0.33 & 0.34 &\it 0.41 \\
\spotexpressive
& 0    & 0.63  & 0.38 &\bf 0.54 & 0.36        &\bf \textit{0.48}\\
\hdashline
\multicolumn{2}{l}{\textit{\;\;\;\;Few-Shot Prompting}}\\
\multirow{3}{*}{\spotexpressive} 
& 3    & 0.64  & 0.37  & 0.50  & 0.34        &\it 0.42\\
 & 6    & \bf 0.67  & \bf 0.39 & 0.51 & 0.35 &\bf \textit{0.48}\\
 & 9    & 0.64 & 0.38 & 0.40 & \bf 0.37      &\it 0.45\\
 
\hdashline
Random Predictor && 0.33 & 0.33 & 0.33 & 0.33& \it 0.33\\
\hdashline

\multicolumn{2}{l}{\textit{\;\;\;\;Cascade Topline}}\\
(ASR)+\llamatwo+(TTS) & 0  & 0.65 & 0.36 & 0.33 & 0.33 &\it 0.42\\
\hline
Prompt Performance  &  & \multicolumn{2}{c}{0.86} & \multicolumn{2}{c}{0.96} \\
\bottomrule
\end{tabular}
}}
\caption{\textbf{Zero-Shot and Few-Shot Performance on the \sentimentbenchmark}. \spot models 
are presented with prompts expressing a positive, negative, or neutral sentiment. In the speech modality, the sentiment comes from vocal characteristics (expressive styles such as sad, laughing, etc.), and in the text, it comes from the semantic content. The continuation is then elicited across modalities or
in the same modality, and tested with pretrained classifiers. 
The last row (Prompt Performance) presents the performance when we apply the classifier directly on the text or speech prompt.}
\label{tab:expr_cont}
\vspace{-1.2em}
\end{table}

\paragraph{Speech and Text Tokenization} We tokenize text with the default LLaMA's tokenizer and speech with the HuBERT tokenizer described above. 
Following previous work, HuBERT tokens are deduplicated for betting modeling quality. For uni-modal datasets (Text-only and Speech-only), we tokenize the data and prepend them with the corresponding modality token, i.e. "\textsc{[Text]}this is a text sentence" or "\textsc{[Speech]}[Hu262][Hu208][Hu499][Hu105]".

\paragraph{Interleaving Speech and Text} For the aligned Speech+Text datasets, we mix text and speech by interleaving speech and text at the word level (Figure \ref{fig:boat1}.b), making the input look like this "\textsc{[Text]}the cat \textsc{[Speech]}[Hu3][Hu7]..[Hu200]\textsc{[Text]}the mat"\footnote{with "[Hu3][Hu7]..[Hu200]" being the tokenization of the spoken utterance "sat on"}. Our hypothesis is that interleaving training will help the model learn an alignment between speech and text, unlocking better text to speech transfer. The speech and text spans within the sentences are sampled randomly at each training step.

\paragraph{Speech Decoder} As for speech synthesis from speech tokens, we train a HifiGAN \citep{kong2020hifigan, polyak2021speechresynthesis} vocoder on the Expresso dataset. The HifiGAN model is conditioned on HuBERT speech tokens and 1-hot speaker embedding from one of 4 Expresso's voices. 
During training, the HifiGAN model receives duplicated tokens but we also train it jointly with a duration prediction module as used in \citealp{lee-etal-2022-direct, lee-etal-2022-textless}\footnote{https://github.com/facebookresearch/speech-resynthesis/tree/main/examples/speech\_to\_speech\_translation}, which takes as input the deduplicated HuBERT tokens and predict their lengths. During inference, the deduplicated tokens are repeated with the corresponding predicted durations, and are feed into the HifiGAN model to produce waveform.


\subsection{\spotexpressive}\label{sec:spotexpressive}
Previous work shows that HuBERT tokens can capture good phonetic information from speech but perform badly at expressivity \citep{nguyen2023expresso}.
Our goal is to have a model that can understand and preserve the emotion in the input speech while being biometric-free.
We therefore supplement phonetic speech tokens from HuBERT with additional \textit{pitch tokens} and \textit{style tokens} and include them in language model training so that our trained \spotexpressive model can capture and generate more expressive speech.


\paragraph{Pitch Tokens} Following \citet{polyak2021speechresynthesis} and \citet{kharitonov2022pgslm}, we produce pitch tokens using a VQ-VAE \citep{vandenoord2017VQVAE} model trained on the F0 of the input speech.
Following the implementation of \citet{polyak2021speechresynthesis},
we trained a VQ-VAE model on the Expresso \citep{nguyen2023expresso} dataset with a codebook size of 64 and a downsampling rate of 128, resulting in 12.5 pitch tokens per second.
For training the pitch quantizer, the F0 is extracted using pyaapt\footnote{https://github.com/bjbschmitt/AMFM\_decompy}.
However, for the language model training, we extract F0 using FCPE\footnote{https://github.com/CNChTu/FCPE}, a fast pitch estimator using Transformer, for inference speed.

\paragraph{Style Tokens} We extract speechprop features from \citet{Duquenne2023sonarexp_arxiv}, which capture speech input's expressive style. The features were pooled with average pooling over input segments of 1 second, making one feature every one second. We further remove speaker information from speechprop features by fine-tuning the features to predict the expressive style on the Expresso dataset which serves as a normalization step to obtain the style features. We finally train a k-means clustering on the normalized features of Expresso dataset with 100 units.

\paragraph{Expressive Speech Tokenization} We mix the 3 types of tokens (HuBERT tokens at 25hz, pitch tokens at 12.5hz, style tokens at 1hz) into a single sequence of tokens by sorting the tokens with their corresponding timestamps (Figure \ref{fig:boat1}.c). Similar to \spotbase, we deduplicate HuBERT tokens as well as pitch tokens, making the input sequence look like this: "\textsc{[Speech]}[St10][Pi0][Hu28][Hu22][Pi14][Hu15] [Pi32][Hu78][Hu234][Hu468]"

Apart from the speech tokenization, the training details of \spotexpressive are the same as for \spotbase.

\paragraph{Expressive Speech Decoder} We train a HifiGAN model conditioned on HuBERT tokens, pitch tokens, style tokens and 1-hot speaker embedding from Expresso's voices.
The duration predictor is also trained to predict the durations of the HuBERT tokens. During inference, we align each HuBERT token with the corresponding pitch and style tokens and repeat them accordingly.





\subsection{Training Details}

Our \spot models are trained on a combination of speech, text and aligned speech+text sequences.  We report in Table \ref{tab:data} the amount and sampling proportion of each type of data and list the datasets we use here: 

\paragraph{Text-only datasets} We include a subset of LLaMA \citep{touvron2023llama} training datasets, 
where we exclude datasets that are unrelated to speech, like code, 
totaling 300B text tokens.

\paragraph{Speech-only datasets} We employ open-sourced large-scale speech datasets,
totaling 460K hours of speech or 30B speech tokens.

\begin{table*}
\centering\footnotesize
\setlength{\tabcolsep}{1.5pt}
\begin{tabu}{l cc c  cc c cccc c cccc c c c c c cccc}
\toprule
\multirow{2}{*}{\textbf{Model\;\;\;\;\;\;\;\;\;\;\;\;\;\;Task}} &
\multicolumn{2}{c}{\bf WUGGY$\uparrow$} && \multicolumn{2}{c}{\bf BLIMP$\uparrow$} &&
\multicolumn{4}{c}{\bf Topic-StoryCloze$\uparrow$} && \multicolumn{4}{c}{\bf StoryCloze$\uparrow$} &&
\multicolumn{1}{c}{\bf {MMLU$\uparrow$}}\\
\cline{2-3}\cline{5-6}\cline{8-11}\cline{13-16}\cline{18-18}
& T & S && T & S && T & S & T$\rightarrow$S & S$\rightarrow$T && T & S & T$\rightarrow$S& S$\rightarrow$T && T \\
\midrule
\multicolumn{1}{l}{\textit{\;\;\;\;Previous Work}}\\
GSLM  \citep{gslm}&  $\emptyset$ & 64.8  && $\emptyset$ &  54.2  &&  $\emptyset$  & 66.6  & $\emptyset$ & $\emptyset$ && $\emptyset$ & 53.3 & $\emptyset$ & $\emptyset$   &&$\emptyset$ &&\\

AudioLM \citep{borsos2023audiolm}&  $\emptyset$ & 71.5 &&$\emptyset$ &  \bf 64.7  &&  $\emptyset$ &-- & $\emptyset$ & $\emptyset$ && $\emptyset$  &-- & $\emptyset$ & $\emptyset$   &&$\emptyset$\\

Voxtlm   \citep{maiti2023voxtlm}       & \bf  80.3 & 66.1    && \bf  74.2 &    57.1 &&  --  &--    & -- & -- && --  &  -- & -- & --   && --\\

TWIST \citep{hassid2023textually}&  $\emptyset$ &\bf 74.5 &&   $\emptyset$ &59.2 &&  $\emptyset$  &76.4  & $\emptyset$ & $\emptyset$ && $\emptyset$ &55.4 & $\emptyset$ & $\emptyset$   &&$\emptyset$ \\

\hdashline
\multicolumn{1}{l}{\textit{\;\;\;\;Ours}}\\

\spotbase  &\bf80.3 &   69.0 &&   73.3 & 58.3 &&   \bf98.0 &\bf82.9 &\bf72.7 &\bf88.6 &&\bf   79.4 &\bf61.0 & \bf  59.5 &\bf  64.6 &&\bf   36.9 \\

\expressiveVone~~        &   75.8 &   65.0 &&   73.6 &   54.2 &&   97.9 &   75.4 &   61.6 &   73.2 &&   78.9 &   56.9 &   54.6 &   58.8 &&   33.3 &&\\

\hline

\multicolumn{1}{l}{\textit{\;\;\;\;Cascade Topline}}\\
(ASR +) \llamatwo                   & 84.1& 79.2 && 72.8 & 71.6 && 98.5 &94.76 &  94.76  &  94.76  && 81.9& 75.7 & 75.7  & 75.7  && 46.2 \\

\bottomrule
\end{tabu}
\caption{\textbf{Zero- and few-shot comprehension evaluation}. 
Reporting accuracy based on log-likelihood -- normalized by the number of tokens --  minimization prediction. MMLU is evaluated in the 5-shots prompting setting. The other tasks are evaluated in the zero-shot setting. T refers to the text modality and S to the Speech modality. 
We fill with $\emptyset$ the task and modality that are not supported by the reported system, and with $\_$ the scores that are not publicly available.}
\label{tab:main_table_essentials_w_sentiment}
\end{table*}

\paragraph{Aligned Speech+Text datasets} We use 
a small subset of speech datasets that came along with text transcriptions. We then collect speech-text alignments at word-level either through the provided dataset or by performing an alignment at the word level using the aligner tool from \citet{pratap2023scaling}.
All the alignments are automatically curated, and thus, possible errors in the alignments are admitted.
The speech+text datasets comprise of 110K hours of speech or 7B speech tokens (HuBERT) and 1.5B text tokens. In total, we have 570K hours of speech.
As the number of tokens differs a lot in different modalities, we tuned the sampling weights of the datasets so that the model sees each modality (speech, text, speech+text) roughly equal number of times during training.

\paragraph{Optimization}
We point to the Appendix~\ref{sec:optimization} for extensive optimization details.  


\section{Speech and Text Understanding} \label{sec:speechtextunderstanding}


As illustrated in Table~\ref{tab:generations_tacl}, \spot can generate semantically and expressively consistent content when prompted with speech tokens or text tokens
\footref{demopage}
. In this section, we assess notably the semantic ability of \spot in both single- and cross- modal scenarios by evaluating quantitatively a collection of benchmarks that require generating text or speech tokens, we'll study the \spot expressivity evaluation in Section~\ref{sec:sentimentmodeling}.

\subsection{Evaluation Benchmarks}

\paragraph{Speech- and Text- only Tasks}

We use sWUGGY, sBLIMP, StoryCloze as speech tasks. 
All these tasks probe model's comprehension
by providing different input sequences (hypotheses), one of which is correct, and assessing if the model assigns higher log-likelihood to the correct hypothesis among multiple choices.
We point to \citet{nguyen2020zero} for detailed description of sWUGGY and sBLIMP. Briefly, sWUGGY measures the lexical knowledge of the model and  BLIMP measures the grammatical knowledge of the model.
For WUGGY, we report the accuracy on the combination of in-vocab and OOV subsets.
Given the beginning of a short spoken story, StoryCloze measures 
the high-level semantic understanding and common sense \citep{mostafazadeh-etal-2017-lsdsem} of the model. We use the spoken version of the original storycloze (S-StoryCloze) and the topic-Storycloze (T-StoryCloze) assembled by \citet{hassid2023textually} based on simpler negative samples. All of these tasks have a random baseline performance of 50\% and are evaluated in the zero-shot setting. 
In addition to speech, these benchmarks are also available in the text modality. We, therefore, measure the text-modeling abilities of \spot on these. In addition, we evaluate \spot on MMLU \citep{hendrycks2021mmlu}, a popular evaluation benchmark for text-based LLMs, using a 5-shot setting.
All the tasks are reported with the accuracy metrics.

\paragraph{Cross-modal Speech-to-Text and Text-to-Speech Tasks}
\spot is trained in both speech and text. For this reason, it has the ability to model tasks that require both text and speech modeling. Based on the text and speech versions of StoryCloze, we build speech to text (S$\rightarrow$T) and text to speech (T$\rightarrow$S) Storycloze for which the context is in one modality (e.g. speech) and the hypothesis is in the other modality (e.g. text).
They are evaluated similarly to other comprehension tasks by comparing the log-likelihood given by the model and are performed in the zero-shot setting.
We also evaluate \spot in-context learning capability with few-shot generation tasks: ASR, TTS, and Speech Intent Classification.
For ASR, 
we prompt the model with examples of speech-text transcription pairs along with a new speech segment for the model to generate the text transcriptions.
We report the Word-Error-Rate (WER) between the generated and the gold transcriptions.
For TTS, 
we prompt the model with examples of text-speech pairs and a new text to be synthesized.
We transcribe the generated audio with Whisper \citep{radford2023robust} and compare it with the original text with Character-Error-Rate (CER). Both these tasks are evaluated with Librispeech clean and other test sets.
We use the Intent Classification task from \citet{chang2023speechpromptv2}. Similar to the ASR task, we prompt the model with examples of speech-intent text pairs and a new speech input. We evaluate model generation with the exact match accuracy metrics.
We report the detailed prompting used for few-shot generation tasks in the Appendix~\ref{apx:few_shot_prompt}.





\paragraph{Baselines} We compare our results with previously published generative speech systems.
All these methods use one or several Transformer \citep{vaswani2017attention} decoder-only models trained on speech units. They differ in how they are trained (pretrained from scratch or fine-tuned), the types of speech units they model, and their amount of training data. We compare with 
GSLM \cite{gslm}, TWIST \citep{hassid2023textually} based on Llama-13B and AudioLM \citep{borsos2023audiolm}.
In contrast with \spot, the approaches mentioned above only rely on speech units during training, making them speech-only models (i.e. they  do not support text understanding nor generation). We also compare our models to VoxtLM \citep{maiti2023voxtlm}, a concurrent work on speech and text language modeling.
We report the best scores from the original published papers for all the mentioned methods. As a top-line comparison, we compare our models with cascade models that use \llamatwo as a text generative model. For text-to-text (T$\rightarrow$T), we only rely on \llamatwo -7B. For speech-to-speech (S$\rightarrow$S), we utilize the cascade model, ASR from \whispermedium\cite{radford2023robust}, followed by \llamatwo, synthesized by \mmstts\cite{pratap2023scaling}.

\paragraph{Ablation Experiments} Finally, we ablate the several components of the \spot training recipe. We compare \spotbase to a \llamatwo model continuously pretrained with two \textit{parallel data training} settings. First, the ASR+TTS-only model consists of training with pairs of semantically equivalent sequences of speech and text (e.g. ``\textsc{[Text]} the cat jumped by the window \textsc{[TTS]}[Hu12]..[Hu54]'' or ``\textsc{[Speech]}[Hu12]..[Hu54]\textsc{[ASR]} the cat jumped by the window''\footnote{with ``[Hu12]..[Hu54]'' being the tokenization of the spoken utterance ``the cat jumped by the window''}). Second, the Word-level Transcription model consists of training on sequences of pairs of textual and spoken words (e.g. ``\textsc{[Text]} the \textsc{[Speech]}[Hu12]..[Hu34] \textsc{[Text]} cat \textsc{[Speech]}[Hu454]..[Hu90]...\textsc{[Text]} window \textsc{[Speech]}[Hu15]..[Hu54]''). Additionally, we compare \spotbase to models trained on a single modality (speech or text) and with speech+text but without any interleaving data (cf. noted ``No Interleaving''). 

\begin{table*}[t]
\centering\footnotesize
\resizebox{0.95\textwidth}{!}{
\begin{tabu}{l cc c cc c c }
\toprule
\textbf{Model}\hfill \textbf{Task} & \multicolumn{2}{c}{\bf LS clean (10 shots)}  && \multicolumn{2}{c}{\bf LS other (10 shots)} && \multicolumn{1}{c}{\bf IC (30 shots)}   \\ 
\cline{2-3}\cline{5-6}\cline{8-8}
& ASR$\downarrow$ &TTS$\downarrow$ && ASR$\downarrow$ &TTS$\downarrow$ && $\uparrow$  \\
\midrule

\multicolumn{2}{l}{\textit{\;\;\;\;\spot variants}}\\

\spotbase  &     21.9&  45.5  && 29.2&    43.8&&   71.9 \\
\;\;\;\;+ASR+TTS      &     \bf6.0 & \bf   6.7    && \bf    11.0 &\bf 7.9 &&\bf 75.8\\ 
\expressiveVone & 37.9& 52.0 &&50.0& 53.6 && 66.2 \\ 

\hdashline
\multicolumn{2}{l}{\textit{\;\;\;\;Parallel Data Training}}\\
Word-level transcription                  & 113.2&  85.2 & & 111.6   & 75. 2&& 22.6      \\
ASR+TTS only      & 7.7 &  8.1   &&   11.9& 9.4&& 7.4   \\

\hline
\multicolumn{2}{l}{\textit{\;\;\;\; Cascade Topline}}\\
(\whisper +) \llamatwo  (+MMS TTS)   &  3.7 &  4.0 & &  7.2 &  4.9 &&  89.6 \\
\bottomrule
\end{tabu}
 }
\caption{\textbf{Few-shot tasks.} 
We evaluate \spot models for Automatic Speech Recognition (ASR) and Text-to-Speech (TTS) Evaluation on LibriSpeech (LS) and Intent Classification (IC). 
ASR scores correspond to Word-Error-Rate (\% WER) evaluated in the 10-shots setting with a max context length of 1024. TTS scores correspond to the Character-Error-Rate (\% CER) in the 10-shots setting with a max context length of 2048. IC scores correspond to accuracy in the 30 shots setting.}
\label{tab:asr}
\end{table*}

\subsection{Single-Modality Performance}
We report in Table~\ref{tab:main_table_essentials_w_sentiment} results on comprehension evaluations.
The reported metrics are calculated with the normalization of the log-likelihood as similar to previous work\footnote{We observe that the normalization of the log-likelihood has different impacts on various tasks, but we follow previous work to normalize the log-likelihood in Table~\ref{tab:main_table_essentials_w_sentiment}. Please refer to Table~\ref{tab:appendix_acc_and_acc_token} for a full comparison.}.

We find that \spotbase competes with the baselines for WUGGY, BLIMP, and Storycloze in the speech modality while preserving competitive text performance.
More specifically, \spotbase outperforms the baselines by a large margin on StoryCloze, which requires the most advanced speech semantic abilities compared to the other reported benchmarks. 


\paragraph{Interleaving is critical} We run ablation experiments (cf. Table~\ref{tab:ablation_zero_shot_and_mmlu}) to understand what leads to this performance by controlling for the training budget and ablating a large number of training parameters. We set the training budget at 100k training steps or 100B tokens. 


%

First, fine-tuning the model on speech-only tokens leads to a much lower performance (e.g. more than 6 points difference with \spot on spoken Storycloze). This shows that interleaving training not only helps preserve the text generation abilities of the model but also leads to better speech understanding and generation performance. Second, we find that fine-tuning \llamatwo on parallel data, ---- both with ASR+TTS only training or Word-level transcription training -- leads to lower performance on tasks such as StoryCloze and BLIMP. Notably, 
the performance is more than 10 points lower on cross-modal Topic-StoryCloze (T$\rightarrow$S and S$\rightarrow$T).

 Finally, we measure the importance of the amount of aligned data used for interleaving training in Figure~\ref{fig:percentage_text_speech_small}. We find that the model's performance in speech (T-StoryCloze) steadily increases with the amount of aligned data. 
Based on these experiments, we conclude that interleaving training is the primary factor leading to good-quality speech generation.

Our interpretation of the superiority of interleaving compared to other mixed-modal training setting and speech-only training is the following: 
Interleaving is the only training recipe that generalize what is learnt during \llamatwo pretraining to speech and text tokens. Indeed, interleaving preserves the right-to-left natural causality of the data within each modality and also across modalities allowing the model to learn aligned representation between speech and text units. We present supporting evidence of this alignment in the next section (\S~\ref{sec:cross_modal_perf})

\paragraph{Expressivity comes with a moderate modeling cost}
As shown in Table~\ref{tab:main_table_essentials_w_sentiment}, \expressiveVone performs lower than \spotbase on these tasks, indicating that the expressive speech units lead to moderate lexical, grammatical, and semantic understanding degradation. This suggests that modeling a given raw speech for \expressiveVone is more costly than for \spotbase. Indeed, in contrast with \spotbase, \expressiveVone is based on integrating expressive speech units in the sequence during training, in addition to Hubert-tokens. This leads to extending the token sequence length for a fixed raw input speech. This added complexity leads to a degradation of speech modeling performance.

In the text modality, despite being fine-tuned on billions of speech tokens, \spot still performs decently on MMLU (above 33\%) and degrades by less than 2 points on WUGGY, BLIMP, and StoryCloze compared to \llamatwo.

Finally, on these tasks, the cascade approach (ASR with \textsc{Whsiper} followed by \llamatwo) is above \spot by a large margin. This may be attributed to the high quality of Whisper ASR and the cleanliness of the benchmarks, which makes the speech content more lossless compared to speech tokenization.

\subsection{Cross-Modal Performance}
\label{sec:cross_modal_perf}
\spot can also model sequences that are made of both speech and text tokens. 

\paragraph{Cross-Modal StoryCloze}  As seen in Table~\ref{tab:ablation_zero_shot_and_mmlu}, we find the performance on StoryCloze in the text to speech direction (T$\rightarrow$S) on par with the speech only performance (S). In contrast, the (S$\rightarrow$T) direction is about 5 points 
above the speech performance (S), suggesting that the model performs better at text generation compared to speech generation even when prompted on speech. 
\begin{figure}[h]
  \centering
  \includegraphics[width=0.5\textwidth]{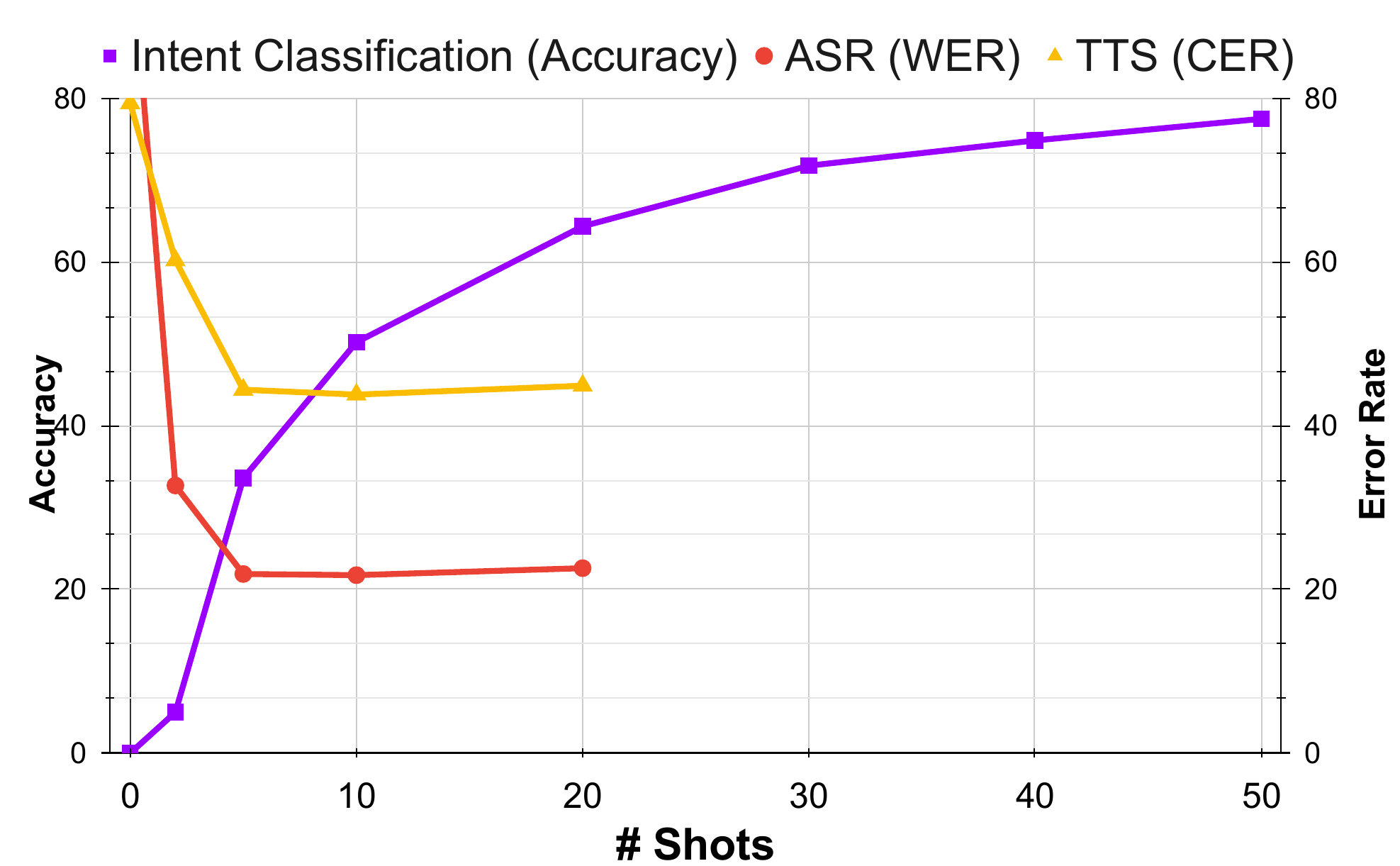}
  \caption{\spotbase performance with regard to the number of shots presented to the model context for Intent Classification, ASR and TTS. }
  \label{fig:few_shots}
\end{figure}

\paragraph{ASR \& TTS}

Similarly to text language models, \spot can be prompted with few-shot examples to perform specific tasks. We illustrate this with ASR and TTS. We show in Table~\ref{tab:asr} that \spot models reach non-trivial performance in ASR and TTS. We find that few-shot prompting leads to the best performance with 10 shot prompting (cf. Figure~\ref{fig:few_shots}).\footnote{We note that above 20 shots, we reach the maximum number of tokens that fit in the context for ASR and TTS.} Our best \spotbase model is at 21.9 WER in Librispeech clean and 45.5 CER in TTS. 
We observe that when we add parallel ASR and TTS examples during training (cf. +ASR+TTS in Table~\ref{tab:asr}), we can improve the performance from a very large margin. We note that adding ASR and TTS data has a very moderate impact on the rest of the tasks. 

\paragraph{Cross-Modal Alignment}

To understand better the hidden mechanism that enables \spot to deliver good cross-modal performance while only being trained on \textit{interleaved} data and raw speech and text, 
we look at the token-level similarity of the model's features from input sequences of HuBERT tokens and the corresponding BPE tokens. 
We illustrate this in Figure~\ref{fig:alignement} (bottom), where we compute the maximum similarity over the same words of speech and text features extracted from different layers of \spot. We find that the similarity between spoken and written sequences inside the model increases from layer 2 and layer 20. In comparison, this alignment is absent in the model trained without interleaving, and is less effective in the model trained with Word-level transcription, particularly in early to middle layers. This suggests that modality mixing enables speech-text alignment, and interleaving further enables the model to map speech sequences with corresponding text sequences.
Figure~\ref{fig:alignement} (top) shows the alignments of BPE tokens and HuBERT tokens of a same sentence.
We see that the middle layers of \spot capture the same semantics information from both input modalities, with high alignments towards the end of each word (last BPE tokens, late HuBERT tokens).

\paragraph{Downstream Speech Classification}
Finally, we report in  Table~\ref{tab:asr} the abilities of \spot to perform Speech Intent Classification (IC) task. We find that the accuracy improves with the number of shots (cf. Figure~\ref{fig:few_shots}). Our best \spot model reaches up to 79\% accuracy (compared to 89\% of the topline performance). 

\paragraph{Pretrained Knowledge is Essential for Few-Shot Learning}

Figure~\ref{fig:learning_curve_small} in Appendix~\ref{apx:complt_res} reports the task-specific performance of \spotbase with regard to the number of training steps compared to a randomly initialized model trained in the same setting. 
After only 25k training steps, \spotbase reaches more than 75\% accuracy on Intent Classification while the randomly initialized model is below 20\%. This means that starting from a pretrained \llamatwo model is essential for few-shot in-context learning and that our method successfully transfers the pretrained few-shot learning abilities of the model to the speech modality.  


\begin{table*}[t]
\centering\footnotesize
\setlength{\tabcolsep}{1.5pt}
\resizebox{\textwidth}{!}{
\begin{tabu}{l cc c  cc c cccc c cccc c c }
\toprule
\multirow{2}{*}{\textbf{Model\;\;\;\;\;\;\;\;\;\;\;\;\;\;\;\;\;Task}} &
\multicolumn{2}{c}{\bf WUGGY$\uparrow$} && \multicolumn{2}{c}{\bf BLIMP$\uparrow$} &&
\multicolumn{4}{c}{\bf Topic-StoryCloze$\uparrow$} && \multicolumn{4}{c}{\bf StoryCloze$\uparrow$} &&
\multicolumn{1}{c}{\bf {MMLU$\uparrow$}} \\
\cline{2-3}\cline{5-6}\cline{8-11}\cline{13-16}\cline{18-18}
& T & S && T & S && T & S & T$\rightarrow$S & S$\rightarrow$T && T & S & T$\rightarrow$S& S$\rightarrow$T && T \\
\midrule
\multicolumn{2}{l}{\textit{\;\;\;\;\spot{} variants}}\\

\spotbase &\bf80.3 &   69.0 &&   73.3 &58.3 &&   98.0 &\bf82.9 &\bf72.7 & \bf 88.6 &&   \bf 79.4 & \bf 61.0 &   59.5 &   64.6 &&   36.9 \\ 

~-~ No Interleaving & 74.7 & 67.1 && 72.6 & 57.2 && 97.7 & 74.0 & 57.5 & 71.9 && 78.2 & 60.1 & 54.2 & 56.4 && 32.1\\  

~-~ Randomly-initialize & 78.1 & \bf 69.9 && 72.9 & \bf 58.8  & & 97.6 & 81.8 & 70.2 & 88.1 & & 73.7 & 58.0 & 58.2 & 62.5 & & 25.8\\   
~-~ Rope $\theta$ default &   78.2 &   69.5 &&   73.3 &   57.7 &&  \bf 98.2 &   82.0 &   72.0 &   88.3 &&   78.9 &   60.9 & \bf 59.8 & \bf 65.5 &&   34.3 \\
 ~-~ +ASR+TTS                   &   76.8 &   68.7 &&   71.7 &   57.2 &&   97.7 &   81.6 &   71.6 &   86.1 &&   77.4 &   59.9 &   58.8 &   63.5 &&   31.4 \\

\hdashline

\multicolumn{1}{l}{\textit{\;\;\;\;Parallel Data Training}}\\

Word-level transcription  &   74.7 &   67.1 &&   72.6 &   57.2 &&   98.0 &   80.3 &   57.5 &   71.9 &&   78.2 &   60.1 &   54.2 &   56.4 &&   32.1 \\
ASR+TTS-only                             &   76.5 & 69.8 &&   73.3 &   57.6 &&   97.3 &   74.9 &   63.5 &   71.8 &&   76.3 &   54.6 &   53.9 &   54.0 &&   34.4 \\

\hdashline

\multicolumn{2}{l}{\textit{\;\;\;\;Unimodal Models }}\\
Speech Only                            &   67.1 &   69.5 &&   53.7 &   58.0 &&   54.8 &   72.9 &   52.2 &   49.4 &&   53.7 &   54.8 &   52.6 &   49.3 &&   27.2 \\
Text Only                              &   72.6 &   46.8 &&\bf73.9 &   52.6 &&   \bf 98.2 &   51.7 &   47.5 &   51.7 &&   79.0 &   50.2 &   47.3 &   52.1 &&\bf 40.1 \\

\bottomrule
\end{tabu}
}%
\caption{
\textbf{Ablation experiments in Zero- and few-shot comprehension evaluation}. All the models reported are initialized from \llamatwo 7B (except Randomly-initialize one) and are trained for 100k steps. Reporting accuracy based on negative-log-likelihood -- normalized by the number of tokens --  minimization prediction. MMLU is evaluated in the 5-shots prompting setting. The other tasks are evaluated in the zero-shot setting. T refers to the text modality and S to the Speech modality. For a full comparison of unnormalized and normalized scoring accuracy, refer to Table \ref{tab:appendix_acc_and_acc_token}.
}
\label{tab:ablation_zero_shot_and_mmlu}
\vspace{-1.2em}
\end{table*}

\section{Expressivity Modeling} \label{sec:sentimentmodeling}
One of the core contributions of this work is the modeling of expressivity. To measure the expressivity of our model we first evaluate the quality of the introduced pitch and style tokens (\S~\ref{sec:expressivity_token_eval}). Second, we evaluate our \spot models on the newly introduced \sentimentbenchmark (\S~\ref{sec:stspbenchmark}). 

\subsection{Style and Pitch Tokens Evaluation}
\label{sec:expressivity_token_eval}

We model expressive speech by complementing phonetic speech tokens (HuBERT) with Pitch and Style tokens. 
To evaluate the quality of our tokenization, we use the speech resynthesis task from \citet{nguyen2023expresso}.
It measures how well the resynthesized speech is compared with the original audio in terms of preserved content, expressive style, and pitch. Table \ref{tab:expressive_tokenizer_evals_avg} shows the performance of \spotbase and \spotexpressive tokenizers compared to Encodec and Hubert-only baselines.  We see the \spotexpressive tokenizer can capture good expressive style and pitch from the input speech. Additionally, we observe a very large improvement in Style and Pitch resynthesis when we compare \spotbase tokenizer with \spotexpressive.

\subsection{The \sentimentbenchmark (\sentimentbenchmarkSHORT)}
\label{sec:stspbenchmark}
To evaluate how well our \spot models can understand and generate expressive speech and text, we introduce the \sentimentbenchmark\footref{stsppage}. It is made of a collection of speech and text prompts in the positive, negative or neutral sentiment. Given a spoken or written prompt , the task consists in generating a text or speech sequence of tokens that preserves the sentiment of the prompt.

For instance, in the text-to-X direction (T$\rightarrow$T and T$\rightarrow$S), given a written sentence bearing sadness, we check if the spoken generated text/utterance is also sad. On the other hand, the direction speech-to-X (S$\rightarrow$S and S$\rightarrow$T), given a spoken happy-sounding utterance, we check whether the model generates a positively written text or positive utterance.

\subsubsection{Sentiment-Rich 
Prompts} \label{sec:sentimentcontinuation:data}
\paragraph{Speech Prompt} In order to have the read speech of different expressive styles (e.g. \textit{he's done it again} in happy/sad style). We utilize two datasets: 1) \textit{Expressive reading} from \expresso \cite{nguyen2023expresso} consisting of 47 hours of expressive North American English speech where 7 different styles are applied on the same content that does not reflect the emotion being conveyed. We use only the speech from 3 emotions: "happy", "sad" and "default". (we will refer to this dataset as \expressoread) 2) \emov\cite{adigwe2018emov}, composed of emotional speech from 5 different speakers and 2 languages (North American English and Belgian French). We select only the English speech from 3 speakers when the same content is recorded in three different emotions: "Amused", "Angry" and "Neutral".

\paragraph{Text Prompt} In order to have expressive text (e.g. \textit{he's such an amazing player} for positive) as prompt, we transcribe\footnote{We use \whispermedium\cite{radford2023robust}} \textit{improvised dialog} from \expresso for 4 emotions: "happy", "angry", "sad" and "default" to obtain an aligned Speech-Text dataset. Then we filter the samples if the transcription has less than 10 words 
or it has one word appearing more than 10 times. We refer to this aligned dataset by \expressoasr.

\paragraph{Sentiment Mapping} To unify different sets of emotional classes, we associate the emotions "happy"/"Amused", "sad"/"Angry" and "default"/"Neutral" to the "positive", "negative" and "neutral" sentiments.

\paragraph{Data Splits} We split the datasets into train/dev/test subsets for later usage. Table~\ref{tab:stsp} presents a comprehensive statistical overview of the datasets used. For \expressoread, we use the original train/dev/test splits; while for the \emov, we split it randomly into train/dev/test subsets with the ratios of 60/20/20. The \expressoasr dataset is also divided into train/dev/set with the ratios of 60/20/20\footnote{We don't use the original data splits because the amount of data in the dev and test subsets is not enough.}. We use the train and dev subsets to train the sentiment classifiers and the test subset to prompt the \spot models.

\subsubsection{Evaluation Metrics}
For both tasks, we check if the generated utterance has a sentiment that is consistent with the sentiment of the prompt. We assess the sentiment of the produced utterance using sentiment classifiers and report its accuracy. We obtain text and speech sentiment classifiers by fine-tuning pre-trained text and speech models respectively. For the speech classifier, similar to \citet{nguyen2023expresso}, we fine-tune the wav2vec2-base model \citep{baevski2020w2v2}
on the training sets of \expressoread, \expressoasr\footnote{We use only the speech data} and \emov. For the text classifier, we fine-tune the 3-classes sentiment classifier from \citet{hartmann2021} on the transcriptions of the \expressoasr training set. The accuracy for speech-to-X directions is averaged over \expressoread and \emov. We repeat the experiments three times and report the averaged accuracy.
\label{sec:sent_cont_eval}

\subsubsection{Evaluation Settings}
We tune the generation parameters on the dev sets, refer to Appendix~\ref{apx:gen_params} for more details.
\paragraph{Zero-Shot}We prompt \spot using positive, negative or neutral text/speech input from the test sets of the datasets described in section \ref{sec:sentimentcontinuation:data}. Then 1) for S$\rightarrow$S and T$\rightarrow$S, we classify the generated speech with the speech classifier. 2) for T$\rightarrow$T and S$\rightarrow$T, we assess the text continuation with the text classifier.

\paragraph{In-context Few-Shot Learning } We also evaluate \spot in a few-shot setting by constructing a set of few-shot examples (cf. Appendix~\ref{apx:sent_cont_few_shots}) and feed them as the in-context prompt. 



\subsubsection{Results}
We report the results evaluated on the test sets in Table~\ref{tab:expr_cont}. For zero-shot performance, \spotexpressive surpasses \spotbase in all directions, with the exception of T$\rightarrow$T where they perform comparably. Compared to the cascade baseline, \spotexpressive outperforms it over all the directions. In the case of few-shot results, we observe that few-shot is beneficial for all directions except S$\rightarrow$S. For both zero-shot and few-shot, the sentiment continuation is better preserved within the same modality than across different modalities. Among all directions, S$\rightarrow$T scores the lowest. The final row of Table~\ref{tab:expr_cont} also includes an evaluation of performance directly on the input prompt. All prompts receive high scores, suggesting a significant potential for improvement in the preservation of expressivity.

\section{Responsible AI in Speech and Text}\label{sec:responsible}
This section discusses and evaluates responsibility aspects from \spot. SpeechLMs have the potential to bring the same benefits as text-based LMs and potentially increase their reach to low-resource languages that are mainly spoken. 

Quantifying and working on user safety is a key aspect from generative model development. These models can inadvertently generate content that is harmful, offensive, or inappropriate is essential for generative language models \cite{deshpande2023toxicity, touvron2023llama}. While safety is a broad concept, we focus on the specific problem of added toxicity in the generation of the \spot. Inspired by previous studies \cite{communication2023seamlessm4t}, we define added toxicity as a toxicity increase in the generation compared to the initial source utterance.
\subsection{Evaluation}
\paragraph{Data} We use the \holisticbias dataset \cite{smith-etal-2022-im}  and its synthesized speech extension \cite{communication2023seamlessm4t}. This dataset has been shown to trigger toxicity for conditional language models \cite{costa2023toxicity}. 
We utilize it as the prompt for generating text (T$\rightarrow$T) and speech (S$\rightarrow$S), respectively. 
We note that this dataset is designed to trigger verbal toxicity. We leave to future work the evaluation of non-verbal toxic content generation (e.g. toxic  sarcasm). 



 \paragraph{Metrics} Similar to \citet{communication2023seamless}, we use \mutox \cite{mutox} and \etox
 \cite{costa2023toxicity} as our  toxicity classifiers. For speech, we simply run ASR and evaluate toxicity with \etox (we refer to this as \asretox). To compute the added toxicity, we evaluate toxicity both in the input prompt and in the generated output. For \etox and \asretox, added toxicity is defined as "when there are more toxic words found in the generated content than in the prompt". For \mutox, added toxicity is identified when the \mutox scores of the generated content exceed the scores of the prompt by more than 0.7.
\begin{table}[h]
\setlength{\tabcolsep}{2pt}
\centering\footnotesize
\resizebox{0.49\textwidth}{!}{
\begin{tabular}{lcccc}
\toprule
\multirow{2}{*}{Task} & \multicolumn{2}{c}{T$\rightarrow$T} & \multicolumn{2}{c}{S$\rightarrow$S}\\
~ & ETOX$\downarrow$ & MUTOX$\downarrow$ & ASR-ETOX$\downarrow$ & MUTOX$\downarrow$ \\
\hline
\spotbase &          1.19         &       2.69       &   1.06 &  3.75\\
(ASR)+\llamatwo+(TTS)  &  1.22      &   2.63   &1.17& 2.70  \\      
\bottomrule
\end{tabular}
}
\caption{\textbf{Added Toxicity Detection}. The proportion of samples with added toxicity divided by the total number of samples. For the \llamatwo baseline, we use a cascaded pipeline made of \whisper for ASR and MMS for TTS.
}
\label{tab:toxicity}
\end{table}
\subsection{Results}
We report results in Table~\ref{tab:toxicity}. In terms of \etox, both \spot and (ASR) + \llamatwo + (MMS-TTS) have comparable results. When evaluated with \mutox, however, \spot shows higher added toxicity especially in S$\rightarrow$S. This might come from the fact that there exists more toxic contents in our speech training dataset. We leave the mitigation to future work.

Figure \ref{fig:toxicity-dist} shows the distribution of added toxicity in \spot in terms of the 13 demographic axes represented in \holisticbias and how they vary in modality. We observe that \textit{Gender and sex} and \textit{Sexual orientation} tend to generate more added toxicity than the rest of demographic axes, while \textit{ability} and \textit{nationality} tend to be among the ones that generate the least. There is no big difference in distribution across modalities or metrics.

\section{Limitations and Broader Impacts}\label{sec:limitsandimpacts}

\paragraph{Harmful applications} \spot also shares the same risks as its generative model predecessors \cite{touvron2023llama}, such as intentionally harmful applications like fake news and spamming as well as unintentionally harmful ones like unfair or biased results, toxic or untrustworthy generations. These risks can be assessed and mitigated using watermarking e.g \cite{kirchenbauer2023watermark} or existing reinforcement learning from human feedback (RLHF) e.g. \cite{bai2022training}.
In addition to these traditional text risks, \spot, being a speech model, also extends risks associated with this modality with intentionally harmful applications like impersonating a specific speaker by continuing short speech segments while maintaining speaker identity and prosody. Mitigation measures for this risk include similar ones as with text (speech watermarking \cite{communication2023seamless} and RLHF). Similarly to text models, unintentionally harm may arise such as the lack of speaker robustness where the model can generate speech continuations inconsistent with the prompt in terms of accent and dialect only for underrepresented groups in the training data. Among the mitigation strategies, we can include: increasing the variety of the dataset, compensating for bias in representation of different demographics.

\paragraph{Future Work} In this paper, we showed how combining style and pitch tokens with phonetic tokens and continuously pretraining a text language model delivers very promising multimodal semantic abilities while enabling expressive speech generations. However, several architectural and training improvements could further progress in speech generation.

First, training multimodal models remains a challenge. In this work, we observed that despite training on both speech and text, our \spot models do not perform as well as the initial \llamatwo model in text. Refining the training could potentially reduce this gap. Second, we restricted our evaluation to English. 
More investigation is needed to assess the quality and safety of the model in non-English languages. 
Third, we only experimented with 7B models. Scaling our experiments beyond 7B could lead to much better performance. 
Finally, the introduced \spot models are foundational models. This means that more work is needed to make them safe and aligned with user expectations. 

\section{Conclusion}\label{sec:conclusion}
 We introduced \spot{}, a language model based on \llamatwo that can generate both speech and text in a cross-modal manner. We showed that by alternating speech and text in the input sequence during training, the model can  generate the content fluidly by changing from one modality to another. We evaluated our models on a collection of speech and text metrics. We plan to make future improvements both in the area of model capability and in transparency and safety.

\bibliography{references}
\bibliographystyle{acl_natbib}
\appendix

\clearpage

\section*{Appendices}

\section{LM Training Optimization}
\label{sec:optimization}
Following \citet{rubenstein2023audiopalm}, we extend the embeddings of LLaMa vocabulary with new speech tokens and modality tokens. The new tokens' embeddings are initialized randomly. 
We then continue to pre-train the 7B \llamatwo model with the constant final learning rate of $3.0e^{-5}$, a sequence length of 4k (equivalent to 200 seconds of speech only), and a batch size of 4 per GPU. We trained the model on 64 A100 GPUs, making an efficient batch size of 1M tokens, for 200K steps, which took approximately 2 weeks.
Following \citet{xiong2023effective} and \citet{roziere2024code}, we make a small modification to the RoPE positional encoding by increasing the “base frequency” $\theta$ of ROPE from 10,000 to 100,000, which has been shown to benefit long-context modeling. Finally, for the speech-text interleaving sampling strategy, we randomly select the word spans so that each text sequence contains 10-30 words and each speech sequence contains 5-15 words, we do this in order to balance the portion of speech tokens and text tokens in the input sequences\footnote{In our initial experiments, we found that changing the length of word spans has little impact on our evaluation metrics, but we do expect a more detailed analysis of this on longer context metrics in further work.}.
\section{Few-Shot Prompts}
\label{apx:few_shot_prompt}
\paragraph{Speech Recognition (ASR)}\mbox{}\\
For ASR, we prompt the model and add special start and end flags. Indeed, we find that without these flags the model tends to hallucinate after transcripting the input sequence. 

For \spot, we use the following prompting. We find that 10 examples leads to the best performance. We illustrate the prompting of \spot for ASR with a single few-shot example:
\resizebox{0.95\linewidth}{!}{
\begin{minipage}{\linewidth}
    \begin{align*}
        &\text{\textsc{[Speech]} \textit{Speech token sequence}}\\
        &\text{\textsc{[Text]} <START Transcript> \textit{Text transcript} <END>}\\
        &\text{\textsc{[Speech]} \textit{Speech token sequence}}\\
        & \text{\textsc{[Text]}}\\
    \end{align*}
\end{minipage}
}
For the models trained with parrallel ASR data (e.g. \spotbase+ASR+TTS), \textsc{[Speech]} is replaced with the \textsc{[ASR]} special token to trigger the transcription prediction as seen during training.

\paragraph{Text-to-Speech (TTS)}\mbox{}\\
 We find that prompting \spot with 10-shots leads to the best performance in TTS. We illustrate the prompting with  a single example for few-shot learning:
 \resizebox{0.95\linewidth}{!}{
\begin{minipage}{\linewidth}
\begin{align*}
 &\text{\textsc{[Text]} \textit{Input Text} 'stop'}\\
 &\text{\textsc{[Speech]} \textit{Speech token sequence} <speech:STOP>}\\
 &\text{\textsc{[Text]}  \textit{Input Text} 'stop'}\\
 &\text{\textsc{[Speech]}}\\
 \end{align*}
 \end{minipage}
}
 With <speech:STOP>, the spoken utterance ``stop'' tokenized into speech tokens\footnote{For \spotbase, the spoken word ``stop'' is tokenized as [Hu481][Hu149][Hu40][Hu48][Hu315][Hu242] [Hu428][Hu494][Hu75][Hu497][Hu188][Hu388][Hu109] [Hu23][Hu338][Hu23][Hu481]}.
 For models trained with parallel TTS data (e.g. \spotbase+ASR+TTS), the token \textsc{[Speech]} is replaced with \textsc{[TTS]}. 

\paragraph{Intent Classification}\mbox{}\\
For Intent Classification, we illustrate the prompting used in \spotbase with single example for few-shot:
\begin{align*}
 &\text{\textsc{[Speech]} \textit{Speech token sequence} \textsc{[Text]}}\\
 &\text{A:activate lights bedroom}\\
 &\text{\textsc{[Speech]} \textit{Speech token sequence} \textsc{[Text]}}\\
 &\text{A:}
\end{align*}
For both ASR, TTS and Intent Classification, we postprocess the output of the model using the special tokens and beginning/end of sequence flags in order to extract the predicted text or speech sequence.

\section{Construction of Few-Shot examples for Sentiment Continuation}
\label{apx:sent_cont_few_shots}
We use S$\rightarrow$T as an illustration, the identical process is applied to the remaining modality directions.
\begin{enumerate}
\item From the \expressoasr training set, we select only the speech samples where the waveform length exceeds 200,000, dividing each into two equal parts. The speech in the second segment is then transcribed.\footnote{The transcription is done by \whispermedium\cite{radford2023robust}.}
\item We apply the fine-tuned speech classifier and text classifier mentioned in \ref{sec:sent_cont_eval} to the speech of the first segment and the transcription of the second segment, respectively. We retain only those pairs where the sentiment of the transcription in the second segment matches that of the speech in the first segment.
\item At the start of each run, we randomly select 3/6/9 samples from the above subset, ensuring a balanced distribution of samples for each sentiment. These samples are then simply concatenated to form the in-context prompt, which is reused for all subsequent iterations.
\end{enumerate}

\section{Generation Parameters}
\label{apx:gen_params}
In terms of the maximal number of generated tokens, we use 50 for T$\rightarrow$T and S$\rightarrow$T, 200 for T$\rightarrow$S, and 300 for S$\rightarrow$S. We use a temperature of 0.8 and nucleus sampling \citep{Holtzman2020The} with a $top\_p$ of 0.95 for all the directions. All the \spot models reported have been trained for 100k steps.

\section{Statistics of the STSP benchmark}
Table \ref{tab:stsp} represents the statistics of the STSP benchmark datasets.

\begin{table}[t]
\centering
\resizebox{0.49\textwidth}{!}{
\begin{tabular}{lccc}
\toprule
  \multicolumn{4}{c}{The \sentimentbenchmark} \\
\midrule
Prompt origin & \expressoread & \expressoasr & \emov\\
Prompt Type &   Speech & Text& Speech \\

\#Samples  & 1020/60/54 & 1373/479/462 & 1053/351/351\\
\#Speakers & 4 & - & 3 \\
Classes & \multicolumn{3}{c}{Positive(33\%) / Negative(33\%) / Neutral(33\%)}\\
\bottomrule
\end{tabular}
}
\caption{Statistics of the \sentimentbenchmark. (\#Samples indicates the number of samples in each train/dev/test split.)}
\label{tab:stsp}
\end{table}

\section{Model Input/Output Samples}
 Table \ref{tab:generations_tacl} shows the generation samples of \spot.


\clearpage

\section{Complementary Results}
\label{apx:complt_res}

\begin{figure}[ht]
  \centering
  \includegraphics[width=0.5\textwidth]{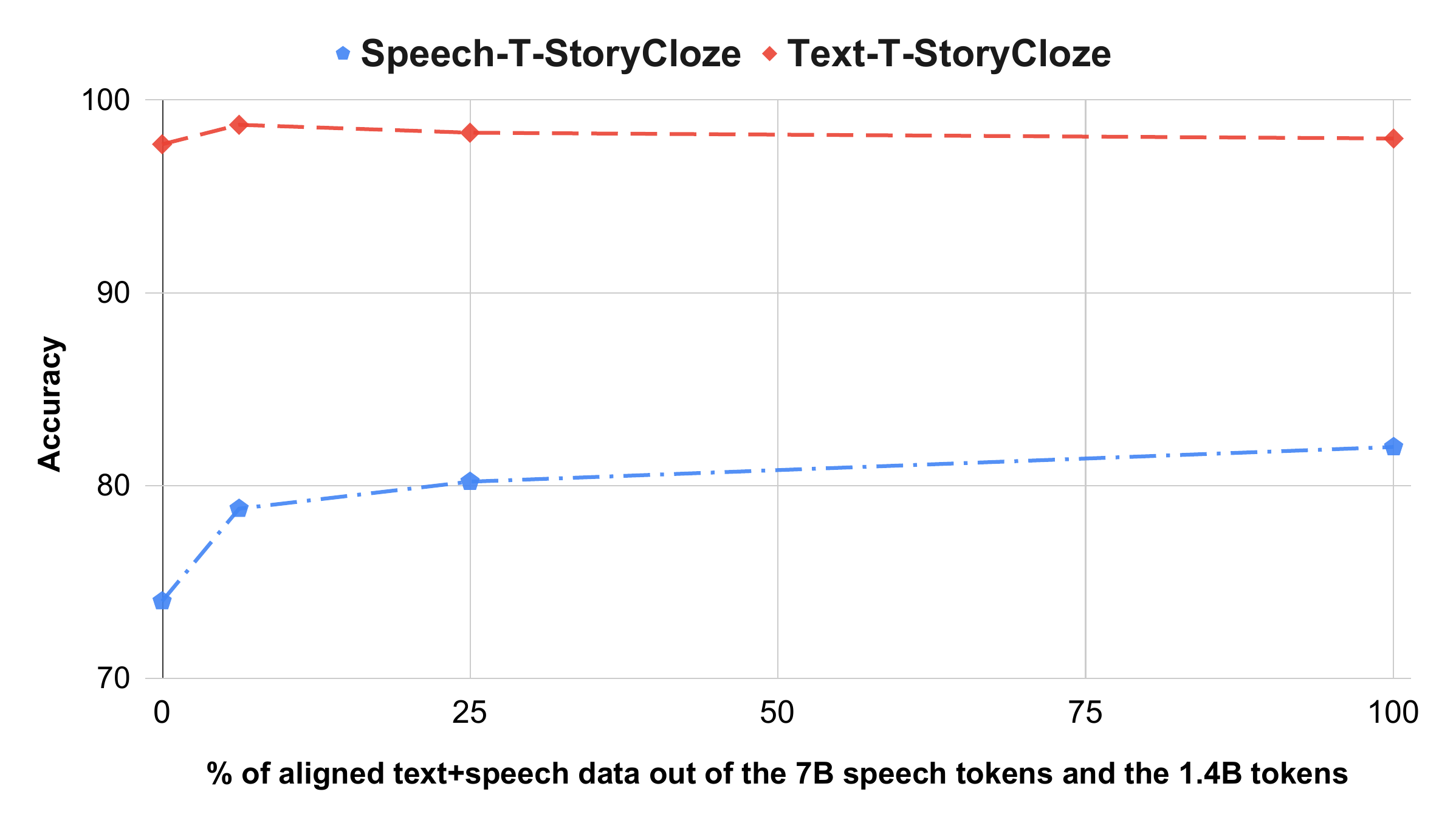}
  \caption{Performance of \spotbase on Topic-StoryCloze in speech and text with regard to the sampled amount of aligned speech+text data from 0\% to 100\% out of the 8.4B aligned tokens (1.4B text and 7B speech). 
  }
  \label{fig:percentage_text_speech_small}
\end{figure}

\begin{table}[h!]
\setlength{\tabcolsep}{2pt}
\centering\footnotesize
\setlength{\tabcolsep}{1.5pt}
\resizebox{0.49\textwidth}{!}{
\begin{tabu}{l cccc}
\toprule
& \bf Bitrate & \bf Content & \bf Style & \bf Pitch \\
\textbf{Model\;\;\;\;\;\;\;\;\;\;\;\;\;\;\;\;\;\;\;\;\;\;\;Metrics } & BPS$\downarrow$ & WER$\downarrow$ & EMO$\uparrow$ & FFE$\downarrow$ \\

\midrule
 \textit{\;\;\;\;Original Audio} & - & 16.2 & 65.2 & - \\

\midrule
\multicolumn{5}{l}{\textit{\;\;\;\;Expresso models \citep{nguyen2023expresso}}}\\
Hubert + HifiGAN & 550 & 23.0 & 22.7 & 0.30  \\
Hubert + HifiGAN w/ GT Style & 550 & 21.4 & 61.6 & 0.27 \\
Encodec (RVQ=1) & 500 & 38.0 & 41.5 & 0.09  \\
Encodec (RVQ=8) & 4000 & 19.0 & 56.7 & 0.04  \\

\midrule
\multicolumn{5}{l}{\textit{\;\;\;\;\spot Tokenizers}}\\
\spotbase & 225 & 23.4 & 20.4 & 0.40  \\
\spotexpressive & 307 & 23.2 & 41.4 & 0.16  \\

\bottomrule
\end{tabu}
}
\caption{\textbf{Expressive Speech Resynthesis Evaluation}. Performances of \spot Tokenizers on the Expresso Benchmark \citep{nguyen2023expresso} compared with their systems. The scores are averaged across datasets. For the detailed scores, refer to Table \ref{tab:expressive_tokenizer_evals_full}.
}
\label{tab:expressive_tokenizer_evals_avg}
\vspace{-1.2em}
\end{table}

\begin{figure}[ht]
    \centering
    \begin{subfigure}[b]{0.5\textwidth}
        \includegraphics[width=\textwidth]{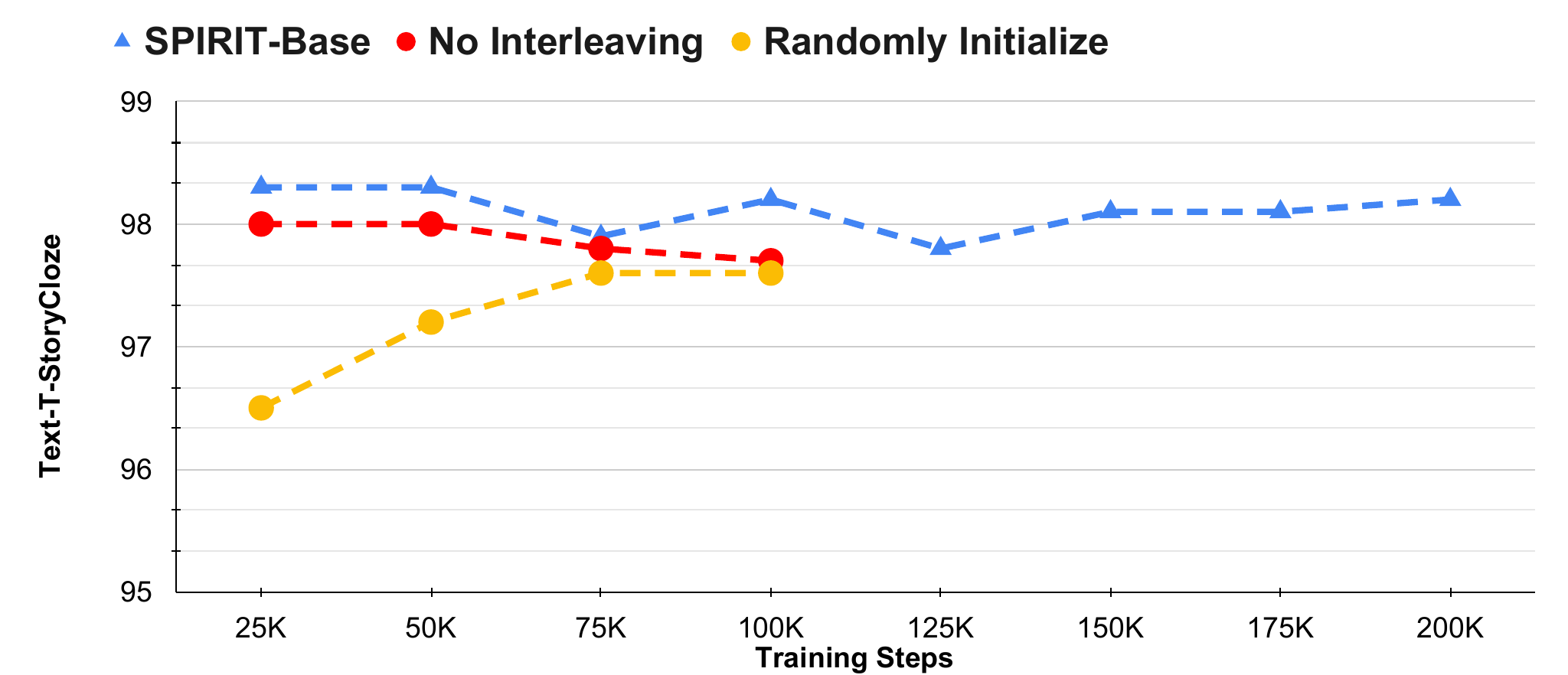}
        \caption{Accuracy on Text T-StoryCloze (0-Shot)}
    \end{subfigure}
    \begin{subfigure}[b]{0.5\textwidth}
        \includegraphics[width=\textwidth]{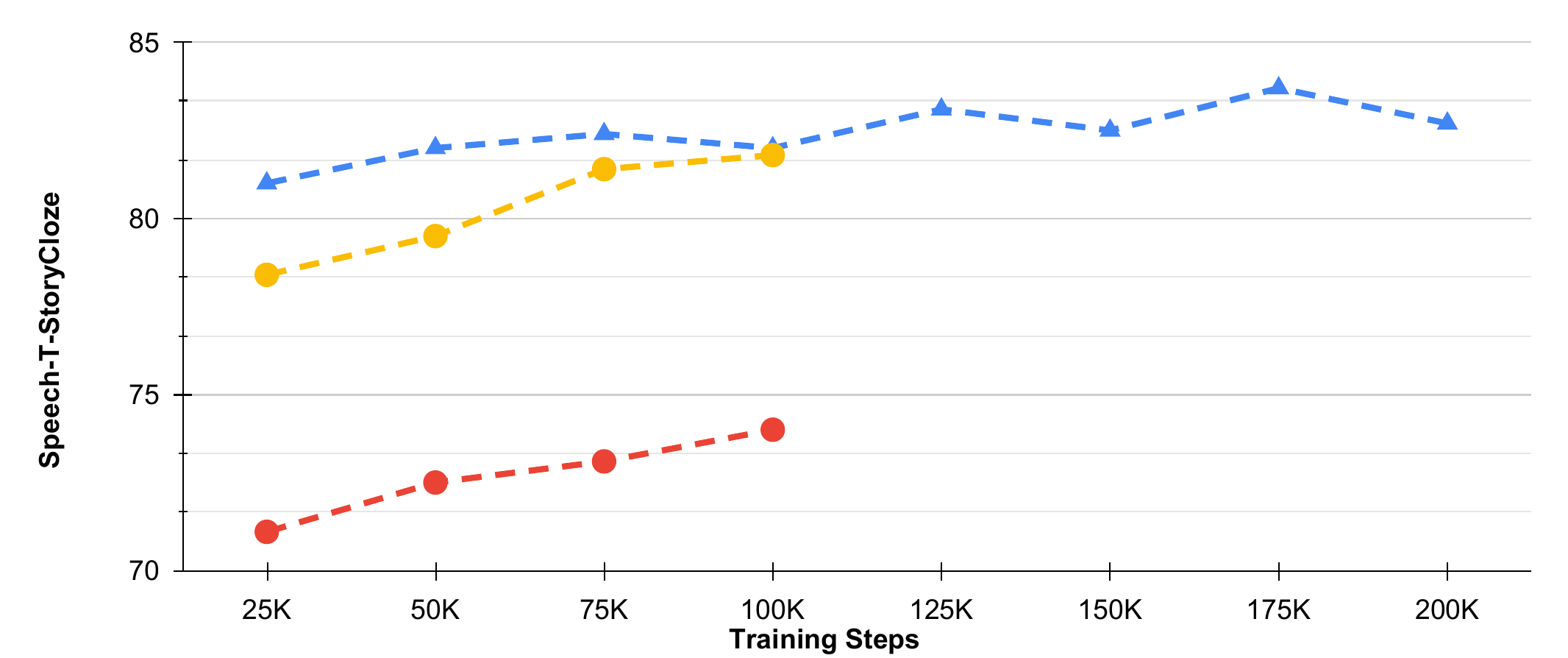}
        \caption{Accuracy on Speech T-StoryCloze (0-Shot)}
    \end{subfigure}
    \begin{subfigure}[b]{0.5\textwidth}
        \includegraphics[width=\textwidth]{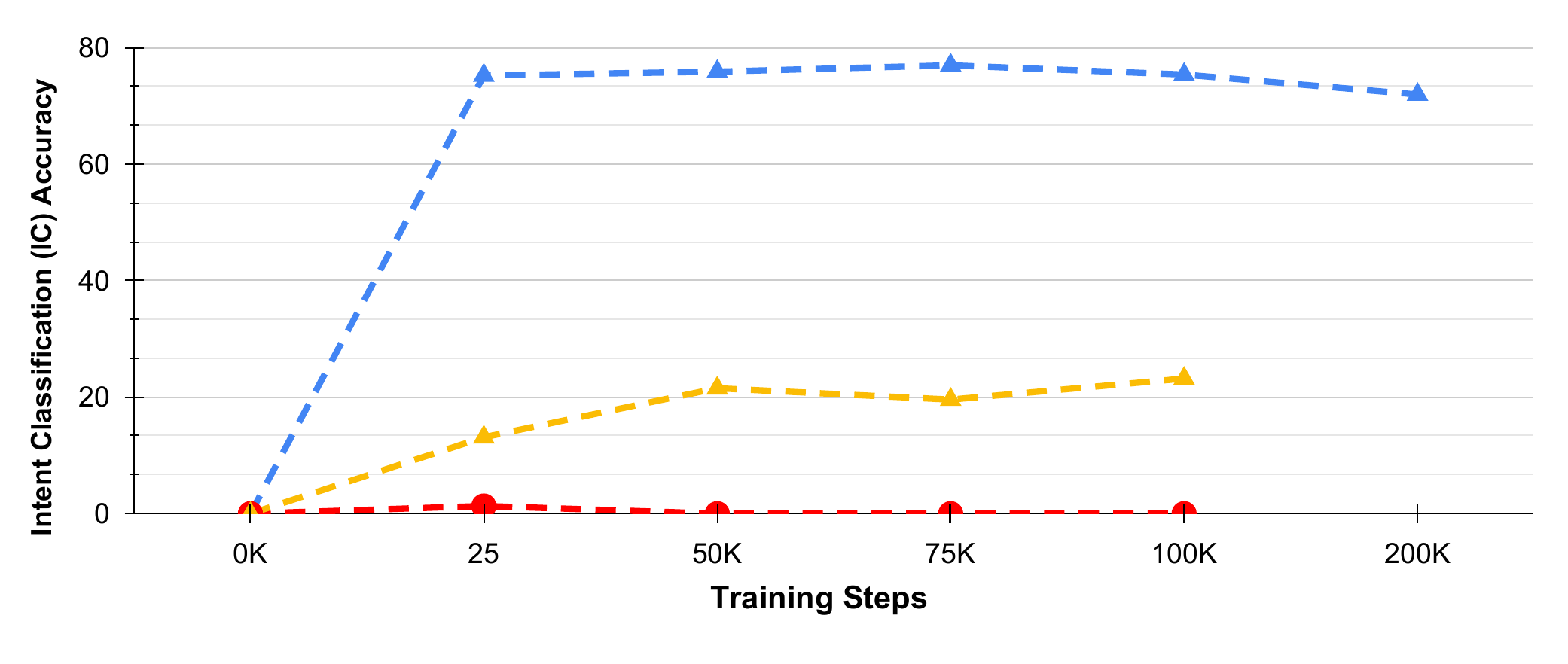}
        \caption{Accuracy on Intent Classification (30-Shot)}
    \end{subfigure}
    \caption{Comparing \spotbase to a randomly initialized model trained in the same way and to a model trained with no Interleaving data. (i.e. the model is only trained on sequences of raw speech or raw text data without any interleaved aligned data.)}
    \label{fig:learning_curve_small}
\end{figure}

\begin{figure}[H]
    \centering
    \includegraphics[width=0.5\textwidth]{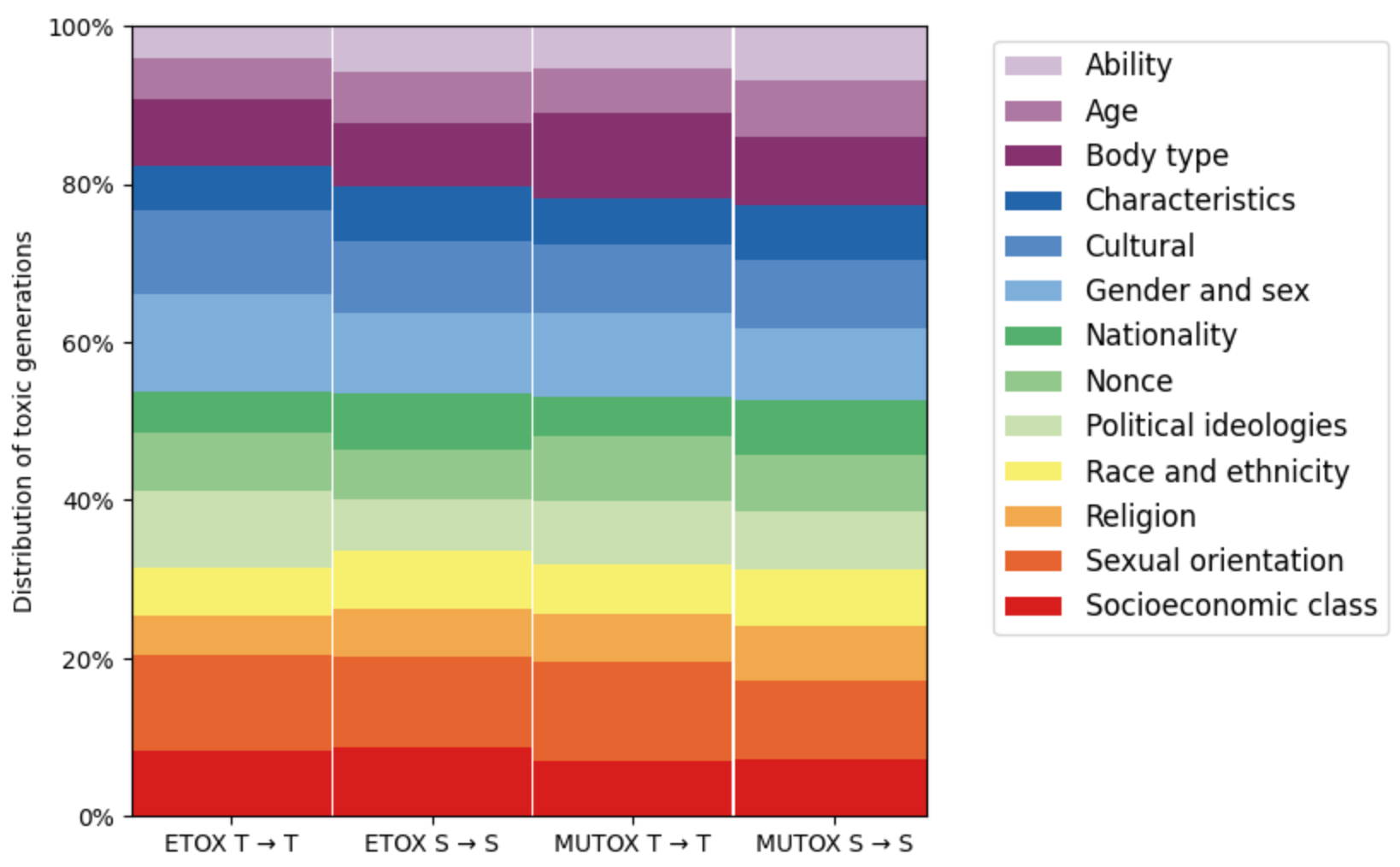}
    \caption{\textbf{Toxicity Distribution} Relative Distribution of added toxicity over the 13 demographic axes for T$\rightarrow$T and S$\rightarrow$S generations. The number of added toxicities are normalized by the number of occurrences in each demographic axis.}
    \label{fig:toxicity-dist}
\end{figure}


\begin{figure*}
    \centering
    \includegraphics[width=\textwidth]{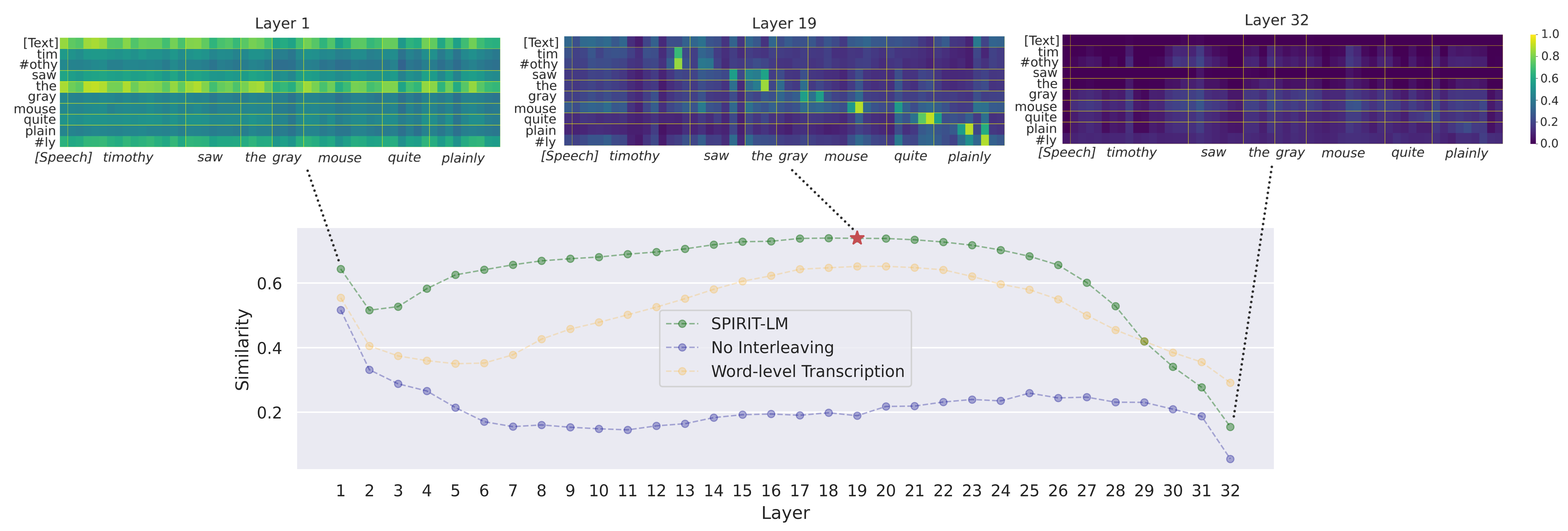}
    \caption{\textbf{Alignments of features obtained from Text and Speech Inputs.} \textbf{Bottom:} Similarity of speech and text features extracted from different layers of \spot compared with the model training without speech-text interleaving. The similarity is computed as the maximum similarity over speech and text features of the same words and is averaged over a test set. \textbf{Top:} Pairwise cosine similarity between
    text features and speech features of the same sentence extracted from different layers of \spot. 
    }
    \label{fig:alignement}
\end{figure*}

\begin{table*}
\centering\footnotesize
\setlength{\tabcolsep}{1.5pt}
\resizebox{\textwidth}{!}{
\begin{tabu}{l cc c  cc c cccc c cccc }
\toprule
\multirow{2}{*}{\textbf{Model\;\;\;\;\;\;\;\;\;\;\;\;\;\;\;\;\;\;\;\;\;Task}} &
\multicolumn{2}{c}{\bf WUGGY$\uparrow$} && \multicolumn{2}{c}{\bf BLIMP$\uparrow$} &&
\multicolumn{4}{c}{\bf Topic-StoryCloze$\uparrow$} && \multicolumn{4}{c}{\bf StoryCloze$\uparrow$} \\
\cline{2-3}\cline{5-6}\cline{8-11}\cline{13-16}
& T & S && T & S && T & S & T$\rightarrow$S & S$\rightarrow$T && T & S & T$\rightarrow$S& S$\rightarrow$T \\
\midrule

\multicolumn{1}{l}{\textit{\;\;\;\;Previous Work}}\\
GSLM  \citep{gslm}&  $\emptyset$ & 65.4/64.8  &&$\emptyset$&  57.2/54.2  &&  $\emptyset$  & 56.3/66.6 &$\emptyset$   & $\emptyset$  && $\emptyset$   & 51.0/53.3 & $\emptyset$  & $\emptyset$    \\

AudioLM  \citep{borsos2023audiolm}&  $\emptyset$ & -- / 71.5 &&$\emptyset$&  \bf -- / 64.7  &&  --  &-- & $\emptyset$   & $\emptyset$  && $\emptyset$   & -- & $\emptyset$  & $\emptyset$ \\

Voxtlm \citep{maiti2023voxtlm}         &  -- / \bf 80.3 & -- / 66.1    &&   -- / \bf 74.2 &    -- / 57.1 &&  --  &--    &--   & --  && --   & -- & $\emptyset$  & $\emptyset$  \\

TWIST \citep{hassid2023textually}&  $\emptyset$ &-- / \bf 74.5 &&   $\emptyset$ & -- / 59.2 &&  --  &-- / 76.4  & $\emptyset$ & $\emptyset$ && $\emptyset$  &-- / 55.4 &$\emptyset$  & $\emptyset$ \\

\midrule
\multicolumn{2}{l}{\textit{\;\;\;\;\spot{} variants}}\\
\spotbase                 & \bf95.1/80.3 &  71.4/69.0   &&  75.7/73.3   & \bf63.2/58.3 &&  94.5/98.0   & \bf69.2/82.9 & \bf66.6/72.7 & \bf83.8/88.6 && \bf76.6/79.4 & \bf56.2/61.0 &  56.2/59.5   &  64.3/64.6   \\
\;\;\;\;+ASR+TTS                   &  94.5/76.8   &  71.8/68.7   &&  74.3/71.7   &  62.4/57.2   &&  93.1/97.7   &  69.1/81.6   &  66.0/71.6   &  81.6/86.1   &&  75.3/77.4   &  55.5/59.9   &  55.5/58.8   &  63.5/63.5   \\

\;\;\;\; Rope $\theta$ default                            &  95.2/78.2   &  71.7/69.5   &&  75.8/73.3   &  62.9/57.7   && \bf94.5/98.2 &  69.5/82.0   &  66.1/72.0   &  83.5/88.3   &&  76.6/78.9   &  56.3/60.9   & \bf56.4/59.8 & \bf64.1/65.5 \\


\spotexpressive         &  95.2/75.8   &  66.2/65.0   &&  76.6/73.6   &  58.7/54.2   &&  94.3/97.9   &  58.2/75.4   &  57.7/61.6   &  81.3/73.2   &&  75.7/78.9   &  51.8/56.9   &  52.5/54.6   &  61.4/58.8   \\

\hdashline
\multicolumn{1}{l}{\textit{\;\;\;\;Parallel Data Training}}\\
Word-level transcription                            &  94.7/74.7   &  71.2/67.1   &&  75.9/72.6   &  62.8/57.2   &&  94.3/98.0   &  68.1/80.3   &  53.9/57.5   &  67.0/71.9   &&  75.8/78.2   &  55.0/60.1   &  51.0/54.2   &  55.1/56.4   \\

ASR+TTS                                &  94.0/76.5   & \bf72.6/69.8 &&  75.7/73.3   &  62.2/57.6   &&  92.7/97.3   &  62.7/74.9   &  56.9/63.5   &  67.8/71.8   &&  73.6/76.3   &  50.7/54.6   &  49.9/53.9   &  53.5/54.0   \\

\hdashline

\multicolumn{1}{l}{\textit{\;\;\;\;Unimodal Ablations }}\\
Speech Only                            &  67.4/67.1   &  71.8/69.5   &&  54.1/53.7   &  63.0/58.0   &&  49.7/54.8   &  62.2/72.9   &  48.3/52.2   &  49.0/49.4   &&  48.2/53.7   &  51.0/54.8   &  48.1/52.6   &  49.2/49.3   \\
Text Only                              &  94.5/72.6   &  53.1/46.8   && \bf77.3/73.9 &  54.6/52.6   && \bf94.5/98.2 &  48.0/51.7   &  47.3/47.5   &  51.5/51.7   &&  76.1/79.0   &  47.0/50.2   &  47.1/47.3   &  50.3/52.1   \\


\midrule

\multicolumn{1}{l}{\textit{\;\;\;\;Cascade Topline}}\\
(\whisper) + \llamatwo                   & -- / 84.1& -- / 79.2 && -- / 72.8 & -- / 71.6 && -- / 98.5 & -- / 94.76 &  -- / 94.76  & -- / 94.76   && -- / 81.9&-- / 75.7 & -- / 75.7  & -- / 75.7 \\

\bottomrule
\end{tabu}
}%
\caption{Zero-shot Comprehension Evaluation in Speech (S) and Text (T). We report Accuracy / Accuracy-token for all the \spot models. 
Both metrics are based on selecting the hypothesis (among two choices) with the highest log-likelihood according to the model. The log-likelihood is based on the sum of each token likelihood in the sequence. The Accuracy is computed based on the prediction that maximizes the log-likelihood of the hypothesis. Accuracy-token adds a normalizing step of the log-likelihood by the number of tokens in the hypothesis. The related work performance (except GSLM) comes from the original published papers of each reported system. We recomputed the scores of GSLM on our metrics.
}
\label{tab:appendix_acc_and_acc_token}
\end{table*}

\begin{table*}
\centering\footnotesize
\setlength{\tabcolsep}{1.5pt}
\resizebox{\textwidth}{!}{
\begin{tabu}{l c ccc c ccc c ccc}
\toprule
& \bf Bitrate & \multicolumn{3}{c}{\bf Content} && \multicolumn{3}{c}{\bf Expressive Style} && \multicolumn{3}{c}{\bf Pitch} \\
\multicolumn{1}{r}{\bf Metrics \;\;\;\; } & \bf BPS & \multicolumn{3}{c}{\bf Word Error Rate (WER)$\downarrow$} && \multicolumn{3}{c}{\bf Classification Accuracy$\uparrow$} && \multicolumn{3}{c}{\bf F0 Frame Error (FFE)$\downarrow$} \\
\cline{3-5} \cline{3-5} \cline{7-9} \cline{11-13}
\textbf{Model\;\;\;\;\;\;\;\;\;\;\;\;\;\;\;\;\;\;\;\;\;\;\;\;\;\;\;\;} & & E. Read & LS & Fisher && E. Read & E. Imp. & EmoV  && E. Read & E. Imp. & EmoV \\
\midrule
\textit{\;\;\;\;Original Audio} & - & 14.76 & 3.55 & 30.26 && 92.47 & 75.69 & 27.46& & - & - & - \\

\midrule
\multicolumn{5}{l}{\textit{\;\;\;\;Expresso models \citep{nguyen2023expresso}}}\\
Hubert + HifiGAN & 550 & 20.64 & 8.46 & 39.84 && 37.02 & 16.62 & 14.45 && 0.31 & 0.32 & 0.26  \\
Hubert + HifiGAN cond. on GT Style & 550& 19.52 & 8.00 & 36.67 && 72.81 & 62.16 & 49.71 && 0.27 & 0.30 & 0.25 \\
Encodec (RVQ=1) & 500& 34.36 & 18.88 & 60.68 && 57.76 & 44.42 & 22.25 && 0.08 & 0.11 & 0.09  \\
Encodec (RVQ=8) & 4000& 16.85 & 4.62 & 35.64 && 78.65 & 64.53 & 26.88 && 0.04 & 0.05 & 0.04  \\

\midrule
\multicolumn{5}{l}{\textit{\;\;\;\;\spot Tokenizers}}\\
\spotbase & 225 & 22.90 & 11.66 & 35.64 && 28.25 & 19.78 & 13.29 && 0.41 & 0.43 & 0.36  \\
\spotexpressive & 307 & 22.35 & 10.60 & 36.58 && 56.02 & 47.66 & 20.52 && 0.16 & 0.17 & 0.16  \\

\bottomrule
\end{tabu}
}
\caption{\textbf{Expressive Speech Resynthesis Evaluation}. Performances of \spot Tokenizers on the Expresso Benchmark \citep{nguyen2023expresso} compared with their Hubert + HifiGAN (with and without conditioning on the Ground Truth Style) and Encodec (with 1 and 8 codebooks) systems on various datasets: Expresso Read section (E. Read), Expresso Improvised section (E. Imp), LibriSpeech dev-other (LS, \citealp{Panayotov2015librispeech}), Fisher \citep{Cieri2004TheFC}, EmoV \citep{adigwe2018emov}. The resynthesis is done with the same input speaker for Expresso subsets and with random Expresso speaker for other datasets. The bitrate is bit-per-second (BPS) computed as \textit{$log_2$(codebook size) $\times$ n tokens per second}.}
\label{tab:expressive_tokenizer_evals_full}
\end{table*}

\end{document}